\documentclass{article}

\usepackage[T2A]{fontenc}
\usepackage[utf8]{inputenc}
\usepackage[english]{babel}
\usepackage{amsmath}
\usepackage{amssymb}
\usepackage{amsthm}
\usepackage{mathrsfs}
\usepackage{dan2e}

\usepackage{subcaption}
\usepackage{graphicx}

\usepackage{algorithmic}
\usepackage[ruled,vlined]{algorithm2e}
\usepackage{hyperref}
\usepackage{booktabs}

\usepackage[
backend=biber,
style=numeric,
sorting=none,
maxnames=99,
giveninits=true,
]{biblatex}
\theoremstyle{definition}
\newtheorem{defi}{Definition}
\theoremstyle{plain}
\newtheorem{remark}{Remark}
\newtheorem{theorem}{Theorem}

\begin{document}

\Volume{}
\Year{2025}
\Pages{}

\udk{}

\title{Loss Barcode: A Topological Measure of Escapability in Loss Landscapes}


\author{S. A. Barannikov\Addressmark[1,2], D. S. Voronkova\Addressmark[1,3], A. Mironenko\Addressmark[1], I.
Trofimov\Addressmark[1], A. Korotin\Addressmark[1,3], G. Sotnikov\Addressmark[1], E. V. Burnaev\Addressmark[1,3]}

\Addresstext[1]{Skoltech, Moscow, Russia}
\Addresstext[2]{CNRS, IMJ, Paris Cité University, France}
\Addresstext[3]{AIRI, Moscow, Russia}


\markboth{S.Barannikov, D.Voronkova, A.Mironenko, I.Trofimov, A.Korotin, G.Sotnikov, E.Burnaev}{Loss Barcode: A Topological Measure of Escapability in Loss Landscapes}

\maketitle

\begin{abstract}
Neural network training is commonly based on SGD. However, the understanding of SGD's ability to converge to good local minima, given the non-convex nature of loss functions and the intricate geometric characteristics of loss landscapes, remains limited. In this paper, we apply topological data analysis methods to loss landscapes to gain insights into the learning process and generalization properties of deep neural networks. We use the loss function topology to relate the local behavior of gradient descent trajectories with the global properties of the loss surface. For this purpose, we define the neural network’s Topological Obstructions score (``TO-score'') with the help of robust topological invariants, barcodes of the loss function, which quantify the escapability of local minima for gradient-based optimization.
Our two principal observations are: 1) the loss barcode of the neural network decreases with increasing depth and width, therefore the topological obstructions to learning diminish; 2) in certain situations there is a connection between the length of minima segments in the loss barcode and the minima’s generalization errors.
Our statements are based on extensive experiments with fully connected, convolutional, and transformer architectures and several datasets including MNIST, FMNIST, CIFAR10, CIFAR100, SVHN, and multilingual OSCAR text dataset.
\end{abstract}

\begin{keywords}
topological data analysis, deep neural networks, loss surface, SGD, persistence barcodes.
\end{keywords}

\section{Introduction}

Neural network (NN) optimization is a daily routine in most areas of deep learning. Despite the conceptual simplicity of gradient-based methods for NN optimization, certain topics in this area remain understudied. First, the optimization algorithms based on stochastic gradient descent (SGD) achieve almost zero loss with high-dimensional deep neural nets (DNN), although the loss functions of DNNs are non-convex, have multiple saddles and local minima. Second, DNNs have good generalization properties, that is, they have essentially no overfitting \cite{goodfellow2016deep,lecun2015deep}.
Also, compared to non-residual networks, models with skip-connections have smoother and more convex loss surfaces \cite{li2018visualizing} accompanied by higher test performance \cite{he2016deep}, which indicates a correlation between the shape of loss landscape and model generalization ability.
In this work, the aim is to gain insight into the above-mentioned properties of neural network optimization by examining loss function landscapes through the means of topological data analysis.

Topological data analysis (TDA) is an actively developing area in machine learning that uses the topological properties of data distributions to obtain information about their internal structure. TDA has found many applications in the field of machine learning, offering novel approaches to data representation, feature extraction, and model interpretation. The principal TDA technique, \textit{persistence barcodes}, enables the extraction of multiscale topological features from complex and high-dimensional datasets, providing robust and invariant representations of the data. These topological features can be integrated into various machine learning problems, including classification, clustering, anomaly detection, segmentation, and representation learning. By capturing the intrinsic structure and connectivity of the data, TDA helps to uncover hidden patterns and relationships that can be overlooked by traditional machine learning methods. Numerous techniques, such as
regularization, batch and layer normalization, designed to improve model generalization and facilitate the training process, modify the topology of loss landscapes \cite{li2018visualizing}. Thus, gaining a comprehensive understanding of these effects through TDA would be highly beneficial.

In Section \ref{minimabadns} we give the definition of the barcode of DNN loss landscapes, which is further referred to as ``loss barcode''. In Section \ref{gradflowsegm}, we propose the algorithm for the computation of the loss barcode and address its computational complexity in Section \ref{sec:comlexity}. Section \ref{sec:to_score} introduces a numerical characteristic of loss landscape complexity based on the loss barcode. In Sections \ref{sec:embed}, \ref{sec:lowering}, we establish and substantiate the loss barcode lowering phenomenon, while Section \ref{sec:explowering} provides an experimental justification. Sections \ref{sec:exp_gener_ability}, \ref{sec:landscape_complex} relate the loss barcode to the model generalization ability and the complexity of the loss landscape. In Section \ref{sec:transformers}, we provide a larger-scale experiment with transformer architecture in the text domain. In Section \ref{sec:RandomPointsOptim}, we demonstrate the scalability and robustness of the loss barcode computation. Finally, Section \ref{sec:optim_lr} provides insight into the optimization process through the lens of the topological invariants explored.

In this work, our key observations are summarized as follows:
\begin{itemize}
    \item Loss barcodes indicate to what extent any given minimum is difficult to escape via gradient-based optimization.
    \item Loss barcodes can be used as a measure of how far the loss landscape is from a convex function loss landscape.
    \item Loss barcodes can indicate the relative quality of minima in terms of their generalization properties.
    \item Loss barcodes and the TO-score give numerical characterization of loss landscape overall topological complexity.
    \item A robust estimation of a loss barcode can be effectively computed for a given set of minima via optimization of the curves connecting the minima.
\end{itemize}

\section{Related Work}

Various aspects of a loss function and its landscape have been thoroughly studied to gain insight into neural architecture search \cite{liu2019darts}, model selection \cite{li2018visualizing}, model pruning ability \cite{zhou2023three}, and optimization \cite{foret2021shaprness-aware} of neural networks.

Several works have studied the connectivity of the two local optima found by stochastic gradient descent optimization \cite{khan2024sok, kim2024exploring}. According to \cite{goodfellow2014qualitatively, draxler2018essentially, garipov2018loss}, while linear interpolation between two local minima typically contains a high loss barrier, these minima can be connected by non-linear curves of low loss \cite{draxler2018essentially, garipov2018loss}. The work \cite{kuditipudi2019explaining} gives an explanation of the phenomenon of mode connectivity while \cite{zhao2020bridging} relates mode connectivity to robustness against adversarial attacks. \cite{gotmare2018using} establishes the broad generality of the mode connectivity framework and \cite{raghavan2020sparsifying} searches for high-performance paths connecting dense and sparse neural networks. The works \cite{entezari2022therole, ainsworth2023git} hypothesize that, after permutations of the network parameters, local minima can be linearly connected without high loss barriers. Furthermore, \cite{fort2019large} detects the existence of low-loss subspaces connecting a set of local minima. The works \cite{skorokhodov2019loss, czarnecki2019deep} demonstrate that the loss surface conceals regions with arbitrary landscape patterns, implicating its diversity and complexity of optimization.

A straightforward way to characterize the properties of the loss function is by using landscape visualizations. Low-dimensional 1D \cite{goodfellow2014qualitatively} and 2D \cite{li2018visualizing} visualizations are widely used to envision an optimization trajectory and sharpness of the obtained local minima \cite{goodfellow2014qualitatively, keskar2016large}, mode connectivity \cite{goodfellow2014qualitatively, smith2017cyclical}, and asymmetric valleys \cite{he2019asymmetric}, establishing the smoothing effect of depth, width, and residual connections on the loss landscape geometry \cite{li2018visualizing, park2022how}. However, 1D and 2D visualizations \cite{li2018visualizing} imply severe dimensionality reduction and require cautious interpretation. The paper \cite{elhamod2023neuro} also claims that 2D visualizations \cite{li2018visualizing} are applicable only in the vicinity of a single model. In an attempt to alleviate these shortcomings, the work \cite{elhamod2023neuro} has proposed an approach to non-linear landscape visualization based on autoencoders.

The properties of minima are also a subject of study. The well-known ``flat minima'' hypothesis \cite{hochreiter1997flat} states that flat minima generalize better. Although confirmed by some studies \cite{keskar2016large, li2018visualizing}, the network could be reparameterized to transform flat minima to sharp minima \cite{dinh2017sharp}. Moreover, \cite{he2019asymmetric} states that there exist many asymmetric directions at a local minimum along which the loss increases at different rates. Recent works \cite{lyu2022understanding, jang2022reparametrization} develop a scale-invariant approach to sharpness and incorporate loss landscape geometry into sharpness-aware optimization which improves generalization \cite{foret2021shaprness-aware, du2022sharpness-aware, liu2022random, mil2022make}.
\cite{zhu2018anisotropic} studies how SGD with added noise can escape local minima. \cite{chiang2022loss} shows that in the low-sample and few-shot regimes even gradient-free optimizers find well generalizing minima.

In recent years, promising results have emerged from topology integrated into machine learning problems resulting in solutions such as the topological layer \cite{kim2020pllay}, topological autoencoders \cite{moor2020topological,trofimov2023learning}, interpretable topological features based on attention maps \cite{kushnareva2021artificial} and scalar fields \cite{trofimov2024scalar, lux2025topograph}, topological perspective in generative modeling \cite{barannikov2021manifold, yang2025topological, gupta2025topodiffusionnet, southern2024curvature}, representation topology divergence and its applications for analysis of neural network representations \cite{barannikov2022representation, tulchinskii2025rtd} and disentanglement \cite{balabin2023disentanglement}. In particular, recent research analyzed the loss landscape through the lens of topological data analysis. \cite{bucarelli2024topological} derived analytical bound for the sum of Betti numbers for the sublevel set of the loss landscape in the case of regularized and unregularized neural networks. \cite{akhtiamov2023connectedness} utilized the Morse theory to explain the phenomenon of mode connectivity. The topological perspective has also been introduced to analyze the loss surface properties of modern architectures and experimental setups \cite{horoi2022exploring, voronkova20231}. In contrast, in this work, we propose a scalable approach to quantify the optimization complexity of the loss landscape based on 0-dimensional topological invariants that cover the most essential geometrical properties of the underlying manifold.

\section{Loss Barcode}

\subsection{How to Quantify Escapability of Local Minima?}\label{minimabadns}

\begin{figure}
\label{twominima_barc}
\centering
\begin{subfigure}[t]{0.55\textwidth}
    \includegraphics[width=\textwidth]{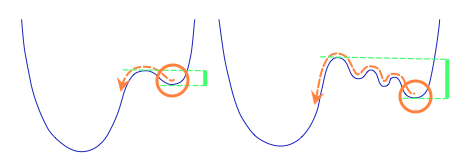}
    \label{fig:twominima}
    \subcaption{}
\end{subfigure}
\begin{subfigure}[t]{0.34\textwidth}
    \includegraphics[width=\textwidth]{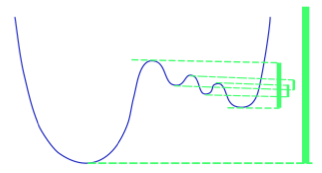}
    \label{fig:simpleBarcodes}
    \subcaption{}
\end{subfigure}
\caption{(a): The two local minima, indicated by circles, look the same locally but pose  different difficulty to gradient-based optimization. The difficulty is quantified by the lengths of the green segments $s_p$, attached to these minima in $\text{Barcode}\,(L)$. $\text{Barcode}\,(L)$ for the simple loss landscape on the right is shown in subfigure (b).}
\label{fig:fig1}
\end{figure}

How can one quantify to what extent a given local minimum represents an obstacle for gradient-based optimization trajectories?
Usually, the escapability of local minima is quantified via the Hessian. However, this is often unsatisfactory. For example, the two local minima in Figure \ref{fig:fig1} pose a different difficulty, although they look the same locally.

The loss barcodes are the numerical invariants that quantify the escapability of local minima, more specifically, the minima's global, reparameterization invariant escapability characteristics. To escape a vicinity of a given local minimum $p \in \Theta \subset \mathbb{R}^d$, a path starting at $p$ must climb the landscape of the loss function $L: \Theta \rightarrow \mathbb{R}_{+}$. However, climbing to the closest ridge point from which the path can then go down in the direction of another minimum does not guarantee that this path descends to a point with loss lower than $L(p)$. For any local minimum $p$ there is the minimum height ($L(p)$ plus ``penalty'') to which a path starting at $p$ must climb, before it can reach a point with loss \emph{lower} than $L(p)$ (see Figure \ref{fig:fig1}).

\begin{defi}
    Consider a path $\gamma: [0,1]\rightarrow \Theta$ that starts at a local minimum $p \in \Theta$ and ends at a point with a loss lower than $L(p)$. Let $m_{\gamma} = \max_t L\big(\gamma(t)\big)$ be the maximum loss value along the path $\gamma$.
    Let $h_p$ be the minimal possible value of $m_\gamma$:
    \begin{equation}h_p=\min_{\substack{\gamma: [0,1]\rightarrow \Theta\\\gamma(0)=p\\L(\gamma(1))<L(p)}}\max_t L\big(\gamma(t)\big) \label{eq:Paths}
    \end{equation} 
    We associate the segment $s_p=[L(p),h_p]$ with the local minimum $p$. The length of the segment $s_p$ represents the \textbf{minimum obligatory penalty} to achieve a lower loss starting from $p$.
\end{defi}
The loss $h_p$ corresponds to a saddle point. The larger the length of the segment $s_p$, the more difficult it is to reach lower loss points starting from a vicinity of $p$ for gradient-based optimization.

\begin{defi}
    \label{def2}
    The \textbf{loss barcode}  is the disjoint union of the segments $s_p$ for all local minima $p \in \Theta$, plus half-line accounting for the global minimum $p_{\text{global}} \in \Theta$: 
    \begin{equation}
    \text{Barcode}\,(L)=[L(p_{\text{global}}),+\infty) \sqcup  \big(\sqcup_p [L(p),h_p]\big) 
    \end{equation}
    See Figure \ref{fig:fig1} for a simple example of a barcode.
\end{defi}

\begin{remark}
The loss barcode from Definition \ref{def2} is equivalent to the $0-$index barcode in the Morse complex.
In Appendix \ref{morse}, we provide a definition of a Morse complex. The definition of a barcode of arbitrary index \cite{B94,viterbo:2011} quantifying ``lifetimes'' of saddles is given in \cite{voronkova20231}. A saddle point of small index and its flat vicinity can also pose obstacles for gradient-based optimization \cite{dauphin2014identifying}. These obstacles are topological in nature, as any critical point or critical manifold.
\end{remark}

\subsection{Calculating Loss Barcodes via Gradient Flow on Segments}\label{gradflowsegm}

In terms of gradient flow, the barcode of the minima can be calculated using the formula (\ref{eq:Paths}) as follows. 

Let $\gamma(t)$, $t\in [0,1]$ be a path from a local minimum $p$  to a lower value minimum $p'$. Consider the action of the gradient flow generated by the gradient vector field $$\dot{\theta}= -\textrm{grad}\,L $$ on the path $\gamma$, up to the reparameterization of the path. At each point on the path $t$ the gradient $\textrm{grad}\,L$ is decomposed into sum $$-\textrm{grad}\,L=(-\textrm{grad}\,L)_{\Vert}+(-\textrm{grad}\,L)_n$$  of two components, where $(-\textrm{grad}\,L)_{\Vert}$ is parallel to the tangent vector to $\gamma(t)$ and $(-\textrm{grad}\,L)_n$ is orthogonal to it.  The action of the tangent component $-(\textrm{grad}\,L)_{\Vert}$ is absorbed into the reparameterization of the path $\gamma(t)$.
Gradient flow moves the curve $\gamma(t)$ by $-(\textrm{grad}\,L)_n$, the normal component of $-\textrm{grad}\,L$.

\begin{algorithm}[ht]
\caption{Barcode of minima computation for loss function of DNN.}\label{algPenlt}
\begin{algorithmic}[1]
\STATE $S_{\text{0}} \gets \{\text{Minima (optimized points)}\, p\in\Theta\}$
\STATE $\text{Barcode}\gets \varnothing$
\STATE {\bfseries for} $p\in S_{0}$ in increasing order of $L(p)$ {\bfseries do}
    \STATE \hskip1em $b[p] \gets L(p)$ 
    \STATE \hskip1em $h[p]\gets +\infty$
    \STATE \hskip1em {\bfseries for} ${(q\in S_{0})} \And{  (L(q)<L(p))}$ {\bfseries do}
     \STATE \hskip2em $\gamma\gets\text{Optimized path} \quad\gamma(0)=p, \quad \gamma(1)=q$
        \STATE \hskip2em $h[p]\gets \min\left(h[p], \max_{t\in [0,1]} L(\gamma(t))\right)$   
    \STATE \hskip1em {\bfseries end for}
    \STATE \hskip1em $\text{Barcode}\gets \text{Barcode}\sqcup \left[ b[p],h[p] \right]$
\STATE {\bfseries end for}
\\
\STATE {\bfseries return} Barcode
\end{algorithmic}
\end{algorithm}

In practice, we can optimize the path by applying the gradient $-(\textrm{grad}\,L)_n$, multiplied by a learning rate, to a set of points approximating $\gamma(t)$. The gradient is calculated via SGD or other gradient-based methods.
The maximum value of the loss function on the set of points approximating $\gamma(t)$ decreases for a sufficiently small learning rate at each iteration. For Lipschitz $L$ and a sufficient number of points, this also ensures that $\max_{t}L\left(\gamma\left(t\right)\right)$ decreases.

\begin{algorithm}[t]
\caption{Optimize path} \label{algo:opt}
\begin{algorithmic}[1]

\STATE {\bfseries procedure} PROJ($\theta_{i}$, $\theta_{j}$):
\STATE \hskip1em \textbf{return} $\frac{\theta_{i} - \theta_{j}}{\|p_i - p_j\|}$
\STATE {\bfseries end procedure}

\STATE

\STATE {\bfseries procedure} STEP(path, $x$):
\STATE \hskip1em {\bfseries for} $i = 1$ \textbf{to} $\lvert$path$\rvert$ - 2 {\bfseries do}
    \STATE \hskip2em $\theta_l \gets \text{path}_{i-1}$
    \STATE \hskip2em $\theta_c \gets \text{path}_{i}$
    \STATE \hskip2em $\theta_r \gets \text{path}_{i+1}$
    \STATE \hskip2em $\nabla f_{\theta_c} = \nabla_{\theta_c} \big( \mathcal{L}(f_{\theta_c}, x) \big)$
    \STATE \hskip1em $\nabla^{left} f_{\theta_c}$ = $\langle \nabla f_{\theta_c}, \text{proj}(\theta_{c}, \theta_{l}) \rangle \cdot \text{proj}(\theta_{c}, \theta_{l})$
    \STATE \hskip1em $\nabla^{right} f_{\theta_c}$ = $\langle \nabla f_{\theta_c}, \text{proj}(\theta_{r}, \theta_{c}) \rangle \cdot \text{proj}(\theta_{r}, \theta_{c})$
    \STATE \hskip1em $\text{path}_{i} \gets$  $\theta_c - \eta \big( \nabla f_{\theta_c} - \frac{\nabla^{left} f_{\theta_c} + \nabla^{right} f_{\theta_c}}{2} \big)$
\STATE \hskip1em {\bfseries end for}
\STATE {\bfseries end procedure}

\end{algorithmic}
\end{algorithm}

To estimate $\text{Barcode}\,(L)$, we repeat the training of a network several times from random initializations to get a sample of minima and then optimize the set of paths between them.
After optimizing a set of paths starting at the given minimum $p$ and going to the points with lower loss, we calculate the segment corresponding to $p$ in $\text{Barcode}\,(L)$ using the formula (\ref{eq:Paths}). This is summarized in Algorithm \ref{algPenlt}.
This algorithm gives a stochastic estimate for $\text{Barcode}\,(L)$.

We experimented with the two variants of the $\gamma(t)$ path optimization procedure. In the first realization (Algorithms \ref{algo:opt}, \ref{algo:insertion}, Appendix \ref{app:insert}),
the location of the points along the segment $t \in [0,1]$ is fixed, and after approximation quality control, if needed, more points are added in between. In the second realization \cite{garipov2018loss}, we take fewer randomly distributed points along the curve. We find that the latter variant is more flexible and has better scalability when the model size increases. Thus, we used it in the experiments in Sections \ref{sec:landscape_complex}, \ref{sec:transformers}, \ref{sec:RandomPointsOptim}.

It should be stressed that, due to the number of parameters and the huge discrete symmetry group action, the number of DNN minima is large and the loss barcode calculation should be considered in the stochastic setting. In particular, given a sample of minima, the result of the above calculation giving a barcode should be considered as a sample from a distribution on the standard space of persistence barcode diagrams, similar to the Vietoris Rips barcodes associated with samples of points from a continuous distribution in Euclidean space as in \cite{chazal2015subsampling}. The experiments demonstrating the scalability and robustness of a stochastic estimation for $\text{Barcode}\,(L)$ are described in Section \ref{sec:RandomPointsOptim}.

\subsection{Computational Complexity}\label{sec:comlexity}

The complexity of the path optimization procedure is linear with respect to $n$, the number of points used for the approximation of each path. The gradient estimation at each point was calculated at each step during the optimization of the paths, which required $O(Cn)$ operations, where $C$ is the number of weights in the neural network. Typically, the time budget for a path optimization is similar to the time budget for a minimum optimization, i.e. single training of a neural network.

The loss barcode computation requires a set of local minima and a set of paths connecting each pair of local minima. Thus, the overall complexity is quadratic with respect to the number of local minima. However, our experiments reveal that the stochastic estimate of the barcode behaves quite stable with increasing number of minima. Hence, in practice, there is no need for large samples of local minima.

\subsection{Topological Obstructions (TO-) Score}\label{sec:to_score}

In the ideal situation, if the loss function has a single minimum and no other local minima, then all gradient trajectories $\dot{x}=-\textrm{grad}\, L$ for all starting points, except for a set of measure zero, always converge to the single minimum.

\begin{defi}
The neural network's \textbf{Topological Obstructions \mbox{(TO-)} score}  is the distance between the minima's barcode of the neural network's loss function $L$ and the minima's barcode of an ideal function $L^{\text{ideal}}$ with the single minimum at the same level as the value of the global minimum $p_{\text{global}}$ of $L$:
\begin{equation}
    \textrm{TO-score}=\textrm{Distance}\left(\textrm{Barcode}\left(L\right),\textrm{Barcode}\left(L^{\textrm{ideal}}\right)\right)\label{TOscore}
\end{equation}
where 
\begin{equation}
\textrm{Barcode}\left(L^{\textrm{ideal}}\right) = (p_{global}, +\infty).
\end{equation}
\end{defi}
The $\textrm{distance}$ here is the standard \textbf{Bottleneck distance} \cite{efrat2001geometry} on barcodes, also known as the Wasserstein$-\infty$ distance $\mathbb{W}_{\infty}$:
\begin{equation}
  \mathbb{W}_{\infty}(\mathcal{D}, \mathcal{D}') := \inf_{\pi\in\Gamma(\mathcal{D},\mathcal{D}')}   \sup_{a\in \mathcal{D}\cup\Delta} 
  \lVert a - \pi(a)\rVert.
  \label{bottleneck_dist}
\end{equation}
Here, the barcode is represented as a point cloud of points $(L(p),h_p)\in\mathbb{R}^2$  and the point $L(p_{\text{global}})\in\mathbb{R}$ on the line ``at infinity'',    $\Delta$ denotes the ``diagonal'' in  $\mathbb{R}^2$   and $\Gamma(\mathcal{D},\mathcal{D}')$ denotes the set of partial matchings between $\mathcal{D}$ and $\mathcal{D}'$ defined as bijections between $\mathcal{D}\cup\Delta$ and $\mathcal{D}'\cup\Delta$.

It can be shown that if $\mathcal{D} = {(b_i, d_i)}_{i\geq 1}$ is the set of (birth, death) points of the corresponding topological features of the loss $L$, then the Bottleneck distance (\ref{bottleneck_dist}) is given by the formula: $\mathbb{W}_{\infty}(\mathcal{D}, \mathcal{D}') = \max_{i, d_i < \infty} \frac{d_i - b_i}{\sqrt{2}}$. 

We provide a general result that relates the zero TO-score and convexity of the underlying function up to reparameterization:

\begin{theorem}\label{th:th1} Let $L$ be a piece-wise smooth continuous function on a domain $D\subset \mathbb{R}^n$, $n\geq 5$, with $-\nabla(L)\vert_{\partial D}$ pointing outside the domain $D$, and such that, for all $r\geq 0$,  index $r$ TO-score$(L)=0$. Then there exists an arbitrary small smooth perturbation of $L$ which, after a smooth reparameterization of the domain $D$, becomes convex.
\end{theorem}

\begin{proof}
See Appendix \ref{app:to_score}.
\end{proof}

As a corollary of the theorem, the smaller the lengths of the segments in the barcode of the loss function $L$, the closer the TO-score to zero, the more similar the loss function $L$ is to be convex up to reparameterization.

\subsection{Loss Preserving Embeddings of Neural Networks}\label{sec:embed}

The neural networks which we consider can be embedded, preserving the loss, into similar networks, which are wider and/or deeper, and which have more parameters. For example, in a fully connected network, one can add an extra neuron in any layer and set all its incoming and outcoming weights to zero. Similar embeddings can be constructed with convolutional architectures.
These loss-preserving embeddings permit one to compare the barcodes of networks of similar type with a different number of parameters. 

Here we describe following \cite{barannikov2019barcodes,fukumizu2000local} the loss-preserving embeddings of fully connected and convolutional neural networks from Sections \ref{sec:fullyconnected},\ref{sec:convn}.
The embeddings are used for the explanation of the loss barcode lowering phenomena described in the next subsection.

A fully connected neural network containing a hidden layer with $k$ neurons with ReLU-activations is embedded in the fully connected network with an additional layer of $k$ neurons in the following way. We place the extra layer directly after the given layer, set the outcoming weights of the extra layer to coincide with the outcoming weights of the given layer, set the weight matrix between these two layers to be the matrix with $1$ on the diagonal and $0$ otherwise, and set the biases of the extra layer to zero. The embedding of parameter spaces $\Theta \subset \Theta'$ defined by this construction preserves the output of the network and therefore preserves the loss. We used the $L_2-$normalization and to compensate for the increases of the $L_2-$norm by $k$, corresponding to the $k$ extra entries of $1$ in the weight matrix, when comparing the barcodes for embedded networks, we shifted the $L2-$norm by the number of neurons added multiplied by the weight decay.

The convolutional neural networks of Section \ref{sec:convn} are similarly embedded in each other: CNN-1-32 $\subset$ CNN-1-64 $\subset$ CNN-1-128, and CNN-2-32 $\subset$ CNN-2-64, preserving the loss.  

\subsection{Lowering of Loss Barcodes with the Increase of Neural Networks' Depth and Width}\label{sec:lowering}

An increase in the number of parameters leads to the lowering of the value of a given minimum, which can drop lower in new directions. It turns out that in the considered networks, the barcodes of minima also get lower. We refer to this effect as the \textit{``loss barcode lowering''} phenomenon, that is, reducing the lengths of the bars, with an increase in depth and the number of channels for shallow neural networks. Here we describe an explanation for the empirical loss barcode lowering phenomenon of Section \ref{sec:explowering} in a simple particular case. 
 
\begin{theorem}
 Assume that two neural network architectures are such that their parameter spaces are embedded in each other as above $\Theta\subset\Theta'$. Let $p_1\in\Theta$ be the minimum of the smaller network with the penalty $h_p$ as defined in Equation \ref{eq:Paths}. Then a minimum of the larger network $p'\in\Theta'$, naturally corresponding to $p_1\in\Theta$, has a smaller penalty $h_{p'}<h_{p_1}$.
\end{theorem}
 \begin{proof}
 Let the path $\gamma_{12}\subset \Theta$ connect the minimum $p_1$ in the smaller parameter space $\Theta$ with a lower loss minimum $p_2\in\Theta$, so that the maximum loss on $\gamma_{12}$ is the penalty $h_{p_1}$ corresponding to $p_1$ as defined in equation \ref{eq:Paths}. Let $p_1'$, $p_2'$ denote the minima of the larger network in the parameter space $\Theta'$ obtained by anti-gradient flow from the points $p_1$,$p_2$ after their inclusion in the bigger parameter space $\Theta'$. To obtain such antigradient flow trajectory, one might need to apply a random fluctuation as the points $p_1,p_2$ become generally saddle points when embedded in the larger parameter space $\Theta'$, see \cite{fukumizu2000local}.
 The obtained trajectories $\gamma_{p_1p_1'}$ from $p_1$ to $p_1'$,  $\gamma_{p_2p_2'}$ from $p_2$ to $p_2'$, and the path $\gamma_{12}$ give together a path $\hat{\gamma}$ connecting the two minima $p_1'$ and $p_2'$ of the larger network. The loss value decreases along the path $\gamma_{p_1p_1'}$ and the path $\gamma_{p_2p_2'}$. Therefore, optimization of the path $\hat{\gamma}$ in $\Theta'$ has a loss maximum lower than $h_{p_1}$. Let $p'$ denote the minimum with the highest loss between the two minima $p_1'$ and $p_2'$. 
 It follows that the penalty $h_p'$ for the minimum of the larger network is lower than the penalty $h_{p_1}$ for the minimum of the smaller network.
 \end{proof}

There is a thorough investigation in \cite{barannikov2019barcodes} of toy models and non-convex functions where this barcode lowering phenomenon was confirmed for small neural networks with few neurons. Therefore, in this work, we aim to apply the technique to the case of modern deep neural networks with cross-entropy loss function and to verify how well it is able to reflect the loss surface structure.

We demonstrate the phenomenon of barcodes lowering with the increase of network depth and width on the MNIST, FMNIST datasets with fully connected neural networks with 2, 3, 4, 6 and 8 hidden layers in Section \ref{sec:fullyconnected}, and on the CIFAR10 dataset in Section \ref{sec:convn} with convolutional neural networks CNN32-1, \mbox{CNN32-2}, \mbox{CNN32-3}, CNN64-1, \mbox{CNN64-2}, CNN128-1. 

\begin{remark} Although typical current DNN in visual domain can have number of parameters  exceeding the number of dataset points by many orders of magnitude, and thus can in principle have any pair of minima connected by a curve with arbitrary low loss, there are two important aspects that must be stressed in this context. Firstly, if the topological (parameterization independent) properties of a given minimum are bad, then it might take an arbitrary big budget to find such a low loss curve connecting it to another same level or lower minimum. We encountered that this tend to happen for example with the fixed small learning rate minima in some experiments on CIFAR10 dataset. Secondly, on the contrary, the most recent  transformer neural networks in language domain have less parameters than the number of dataset points and their loss landscape is known to have much more complex structure with the loss landscape poorly-connected even for a large embedding dimension; see e.g. \cite{yang2021taxonomizing}, Section D.4.
\end{remark}

\section{Lowering of Loss Barcodes}
\label{sec:explowering}

In this section, we show how the loss barcodes are calculated using the proposed Algorithm \ref{algPenlt} and demonstrate the effect of the loss barcode lowering (see Section \ref{sec:lowering} for more details). We further investigate the loss barcode lowering phenomenon for deep residual networks in Section \ref{sec:landscape_complex}.

\subsection{Fully Connected Neural Networks}
\label{sec:fullyconnected}

\begin{figure}[ht]
\centering
\includegraphics[width=0.49\textwidth]{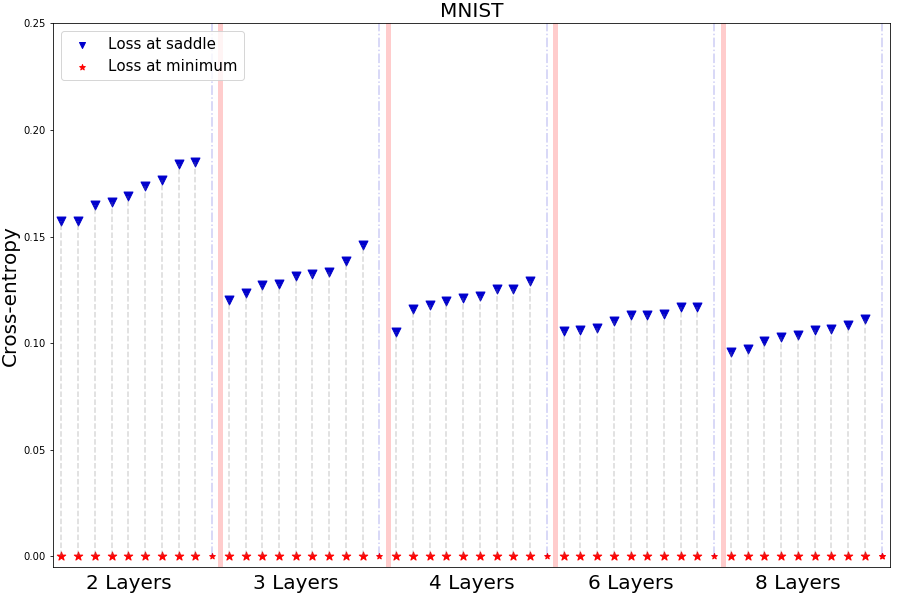}
\includegraphics[width=0.49\textwidth]{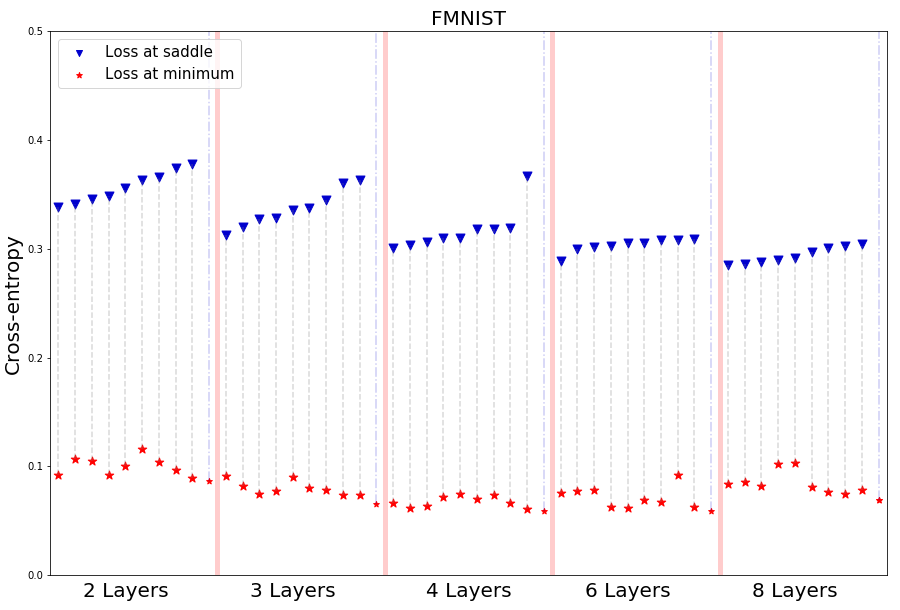}
\caption{Barcodes of fully connected deep neural networks (consisting of 2, 3, 4, 6, 8 layers) trained on MNIST and FMNIST datasets.}
\label{fig:network-barcodesMnist}
\end{figure}

We demonstrate the loss barcode lowering on the MNIST and FMNIST datasets using fully connected neural networks with $2, 3, 4, 6, 8$ layers with $32$ neurons in each layer, see Figure \ref{fig:network-barcodesMnist}. For each architecture, we have obtained $10$ minima by training networks from random initializations until full convergence using Adam \cite{kingma2014adam} and the Cyclical Learning Rate \cite{smith2017cyclical}. Then, we calculated the loss barcodes using the Algorithm \ref{algPenlt} described in Section \ref{gradflowsegm}.

\begin{figure}[ht]
\centering
\begin{subfigure}[t]{.4\textwidth}
    \includegraphics[width=\textwidth]{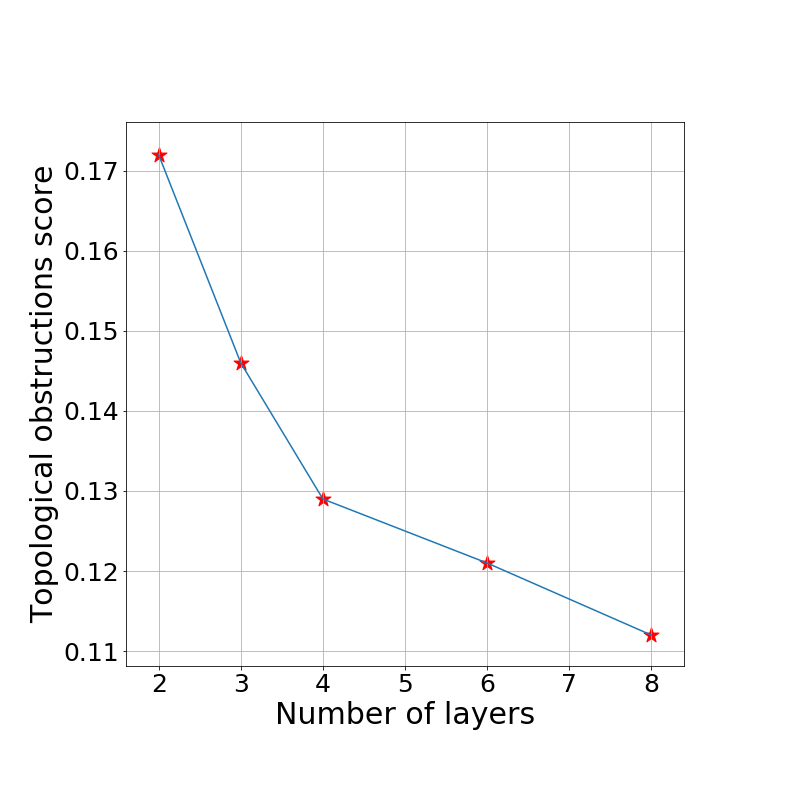}
    \subcaption{MNIST}
\end{subfigure}
\begin{subfigure}[t]{.4\textwidth}
    \includegraphics[width=\textwidth]{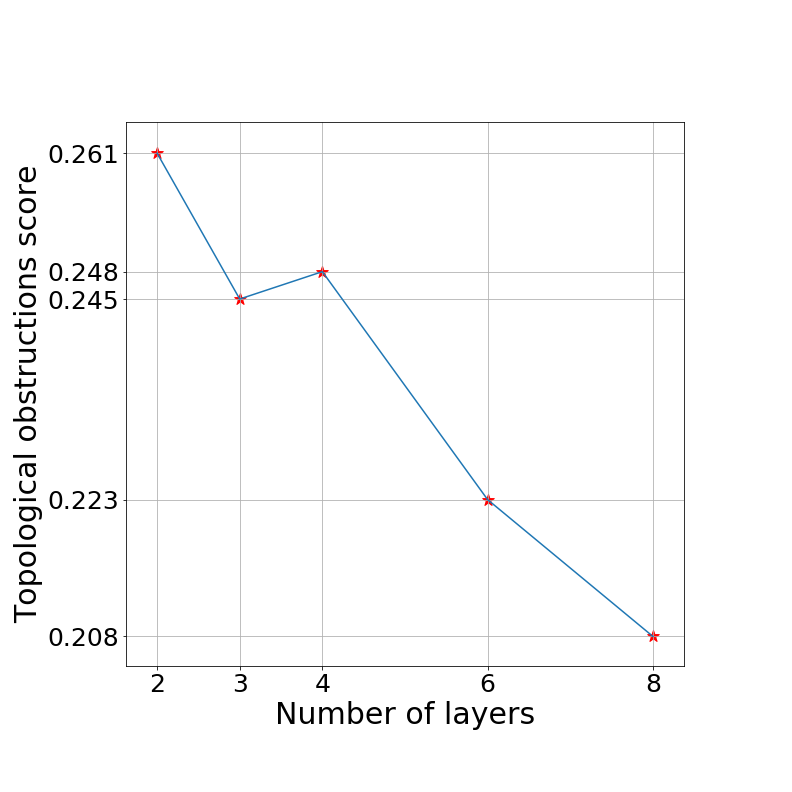}
    \subcaption{FMNIST}
\end{subfigure}
\caption{The effect of decreasing TO-score with the growth of number of layers in FC networks.}
\label{fig:to_score}
\end{figure}

Both the loss barcodes in Figure \ref{fig:network-barcodesMnist} and the corresponding TO-scores in Figure \ref{fig:to_score} reveal that the increase in depth results in a reduction of the lengths of segments in the loss barcodes.

\begin{remark}
Although we define the TO-score as a quantitative characterization of the underlying loss barcode, we mostly provide visualizations of the barcode itself since it reflects more detailed information and, given the loss barcode, the calculation of TO-score is straightforward (see Section \ref{sec:to_score}).
\end{remark}

\subsection{Convolutional Neural Networks (CNNs)}
\label{sec:convn}

To further validate the loss barcode lowering hypothesis, we have experimented with various CNN architectures on the CIFAR10 dataset \cite{Krizhevsky_2009_17719}.  We have used the standard networks, containing one, two and three convolutional layers; see details in the Appendix \ref{app:architectures}. 
The number of parameters in the last FC layer remains constant among CNNs with the same number of channels, to concentrate attention on the convolutional layers. As a downsampling method, the max-pooling operation was used to obtain the same spatial size before the final FC layer for each network. The six convolutional networks from our experiments in this section have between 7K (Conv1-32) and 117K (Conv2-64) parameters.

\begin{figure}[ht]
\centering
{\includegraphics[width=0.7\textwidth]{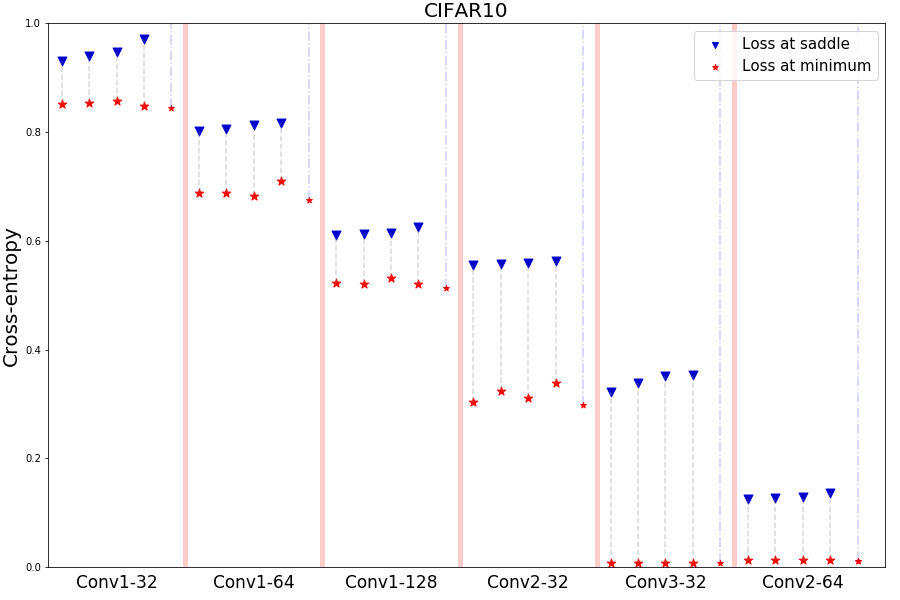}}
\caption{Barcodes of Convolutional Neural Networks on CIFAR10 dataset. Networks sorted in the order of growth of numbers of parameters.}
\label{fig:cnn_all}
\end{figure}

Our experiments presented in Figure \ref{fig:cnn_all} exhibit the loss barcode lowering phenomena. They show the connection between the architecture of a network and the value of the loss function at the saddle points $h_p$ associated with minima. That is, these values decrease with increasing the number of convolutional layers or the number of channels in them. In Figure \ref{fig:cnn_all}, we see a monotonic decrease in the loss function value at the saddle points $h_p$ due to the increase of the number of channels in a single convolutional layer. Moreover, in Figure \ref{fig:cnn_all} one can distinguish even more rapid drop after adding new convolutional layers. Here we sort the architectures by the number of parameters and see the straightforward dependence. However, increasing the number of parameters in a simple architecture might lead to saturation of the effect considered.

\begin{figure}[ht]
\centering
\begin{subfigure}[t]{.49\textwidth}
    \includegraphics[width=\textwidth]{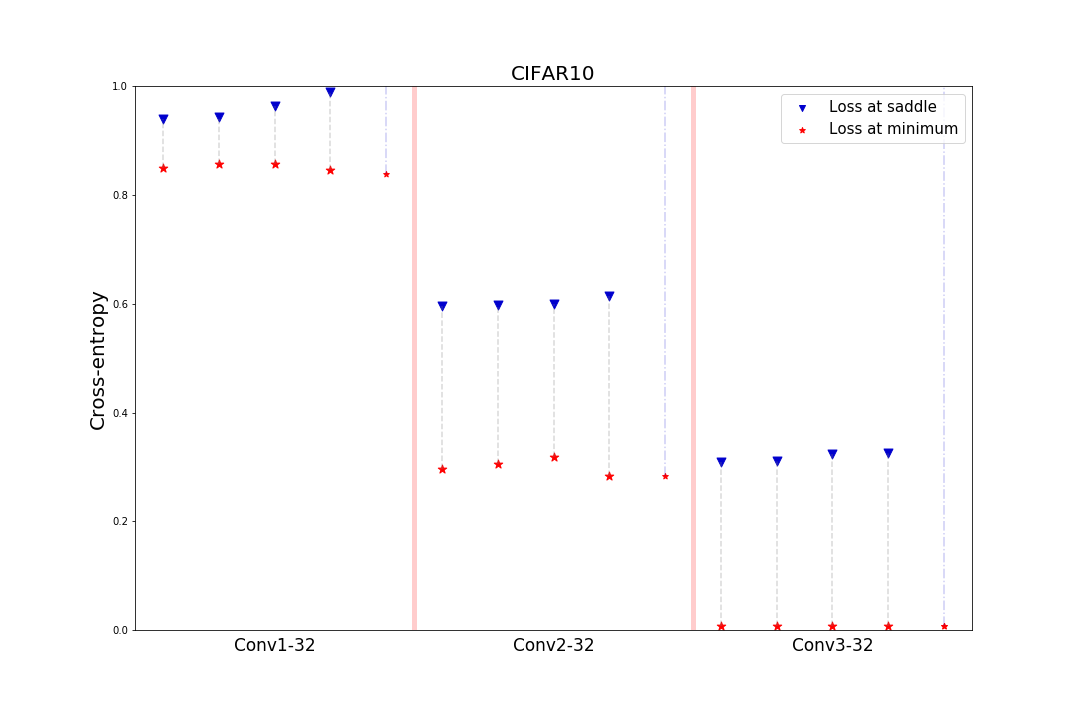}
    \caption{Without Batch Normalization}
\end{subfigure}
\begin{subfigure}[t]{.49\textwidth}
    \includegraphics[width=\textwidth]{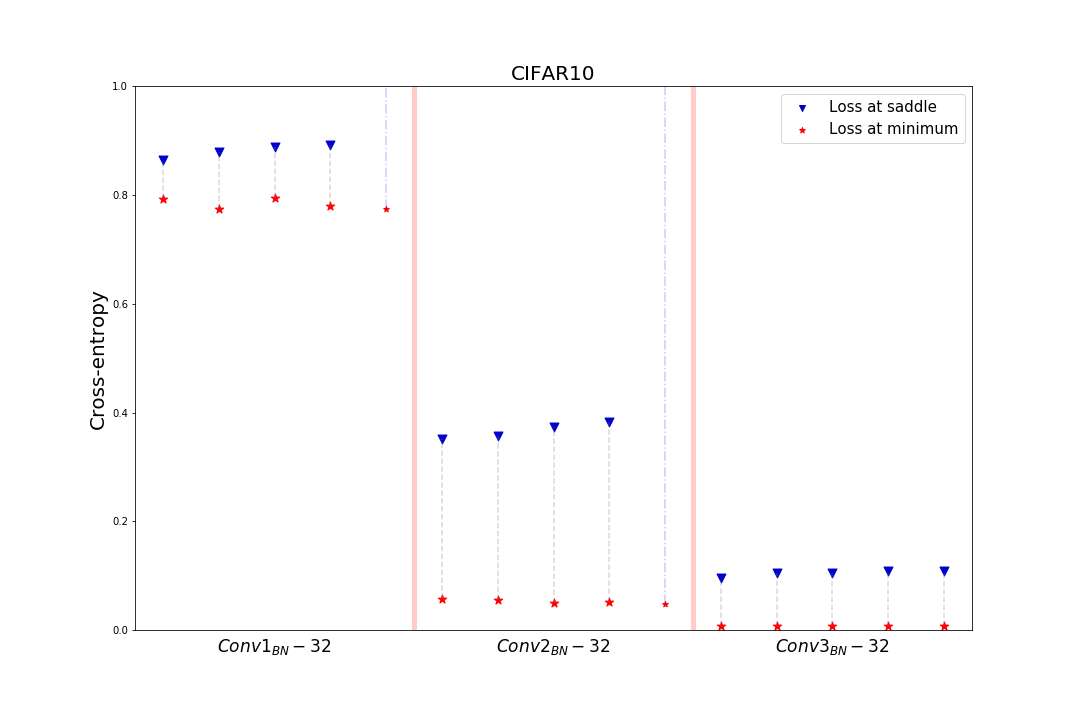}
    \caption{With Batch Normalization}
\end{subfigure}
\caption{Effect of Batch Normalization on the barcodes' height for convolutional neural networks.}
\label{fig:cnn_bn}
\end{figure}

We have also experimented with CNN with batch normalization before the activation function. As mentioned in previous work, batch normalization has a smoothing effect on the loss function; see Figure \ref{fig:cnn_bn}
. We see that among CNNs of all considered depths, their counterparts with BN have both lower loss values in local minima and in saddle points
.

\section{Loss Barcodes and Minima's Generalization Ability}
\label{sec:exp_gener_ability}

In this section, we demonstrate that the lengths of the segments from the loss barcode are
correlated with its generalization ability.

We follow the work \cite{li2019explaining} to investigate the connection between the lengths of the segments in the barcode and the generalization properties of local minima. In the CIFAR10 experiment, \cite{li2019explaining} show that the training procedure with small constant learning rate results in solutions of lower test accuracy compared to those with annealing of the learning rate. This effect is explained by the conjecture that the principal attention is initially attracted by easy-to-generalize, hard-to-fit dataset patterns when the learning rate is a small constant. However, these patterns are initially ignored in favor of easy-to-fit, hard-to-generalize patterns when the learning rate is annealing. To explore the generalization ability in this phenomenon, we analyze the global behavior of loss surface in the neighborhood of local minima learned with both types of training procedures.

For this purpose, we train 8 ResNet-like \cite{he2016deep} models with 195k trainable parameters (the architecture is described in Table \ref{tab:exp_gener_ability:res_arch} in Appendix \ref{app:architectures}) on CIFAR10 and SVHN datasets in two scenarios: using a small constant (below referred to as ``type $1$'') learning rate ($10^{-4}$ for both CIFAR10, SVHN) and applying annealing (below referred to as ``type $2$'') of learning rate (starting from $10^{-2}$ for CIFAR10 and $5\times10^{-2}$ for SVHN and ending with $10^{-4}$ for both CIFAR10, SVHN). To compute the loss barcodes using the Algorithm \ref{algPenlt} in Section \ref{gradflowsegm}, we optimized $28$ paths between the pairs of minima of type $1$ and $28$ paths between the pairs of minima of type $2$
(for further details, see the Appendix \ref{app:train_hyp}).

\begin{figure}[ht]
\centering
\begin{subfigure}[t]{.49\textwidth}
    \centering
    \includegraphics[width=\textwidth]{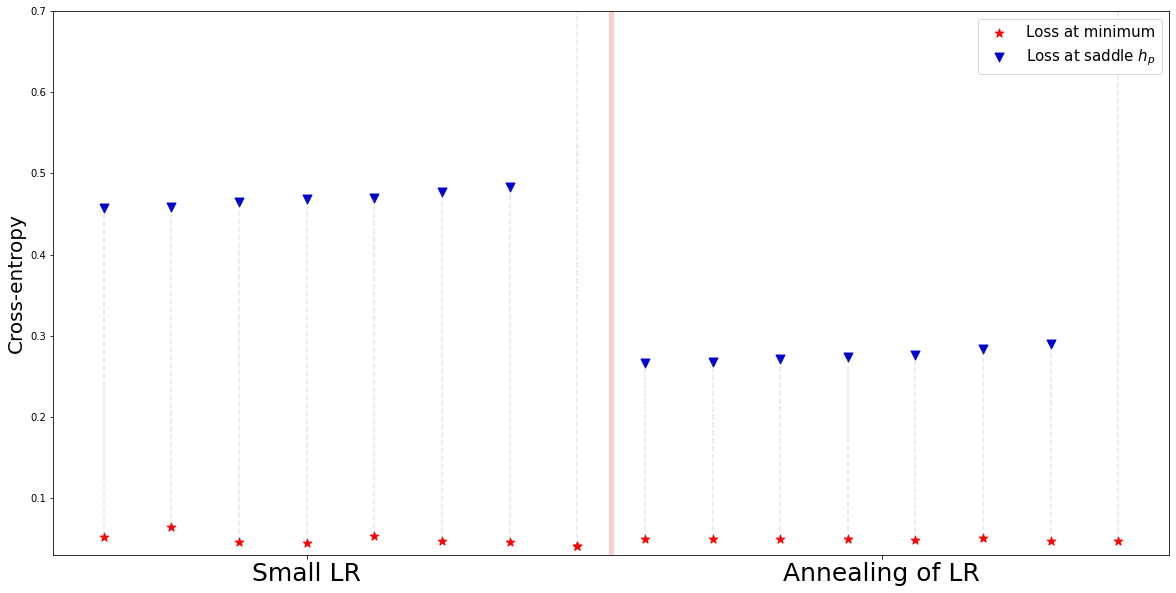}
    \caption{CIFAR10, test acc. (1): $72.66\pm0.83$, (2): $80.64\pm0.41$}
\end{subfigure}
\begin{subfigure}[t]{.49\textwidth}
    \centering
    \includegraphics[width=\textwidth]{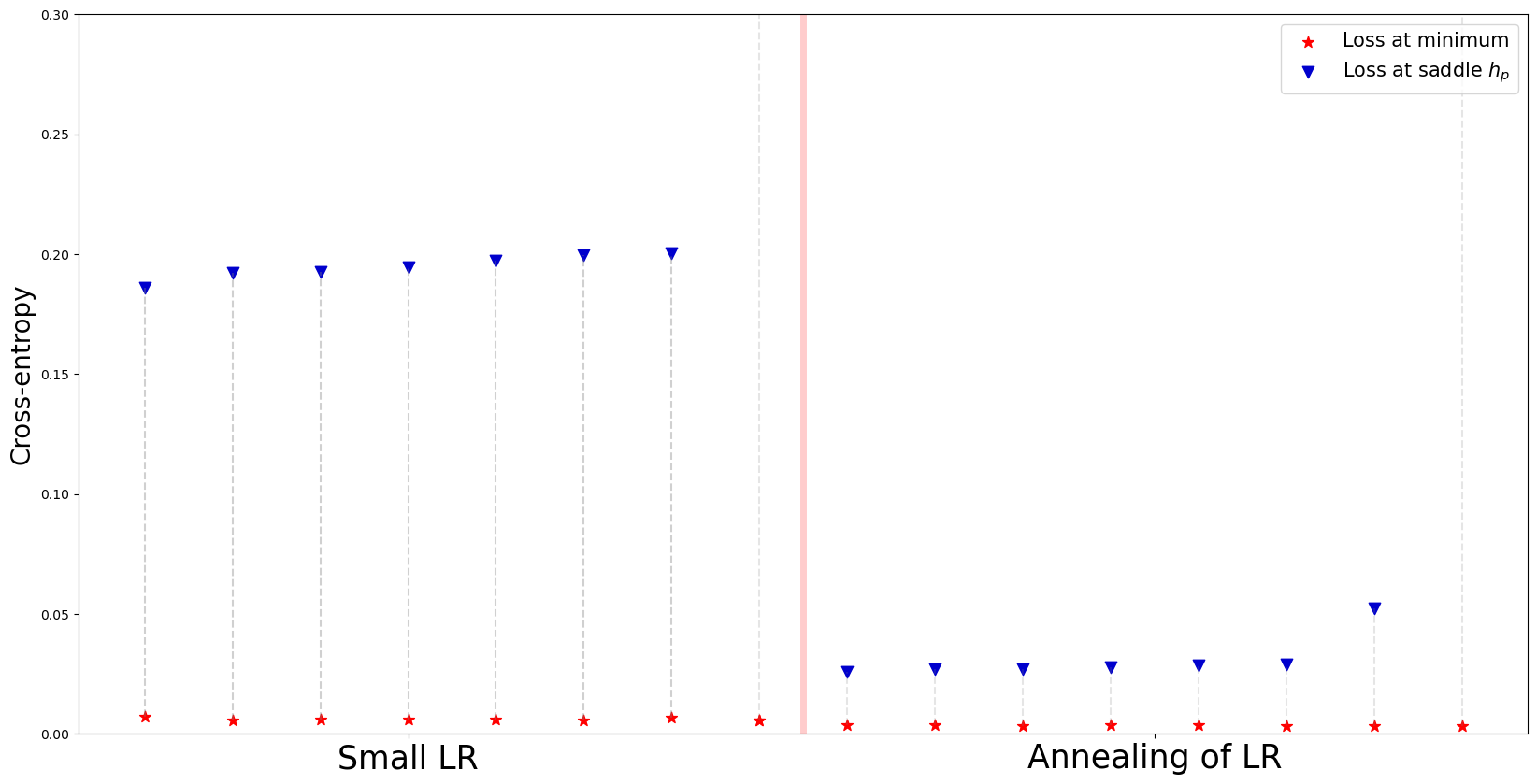}
    \caption{SVHN, test acc (1): $89.99\pm0.40$, (2): $94.22\pm0.19$}
\end{subfigure}

\caption{Differences of barcodes segments length for local minima trained with a constant small learning rate (left from the red line), and with annealing of learning rate (right from the red line). ResNet neural networks.}
\label{fig:exp_gener_ability}
\vskip-0.12in
\end{figure}

For each dataset, we obtain two groups of minima with similar, nearly zero values of train loss but significantly different mean test accuracies: $72.66\%$ (type $1$) vs $80.64\%$ (type $2$) for CIFAR10, $89.99\%$ (type $1$) vs $94.22\%$ (type $2$) for SVHN (for more details, see Table \ref{tab:exp_gener_ability:char_mins} in Appendix \ref{app:exp_details}). There is a sufficient generalization gap between the minima of type $1$ compared to the type $2$ that is captured with the computed barcodes, see Figure \ref{fig:exp_gener_ability}. We observe that loss at the saddle points, or the upper ends of the bars, for minima of type $1$ is always greater than for the minima of type $2$. This observation holds for both the CIFAR10 and SVHN datasets. The experiment reveals that even if models have nearly the same training loss, their generalization potential can differ and can be quantified, using only the training set, by calculating the loss barcodes. Thus, one can make a decision to use a model with a lower loss barcode, as it indicates better generalization performance.

\section{Loss Barcode as a Measure of Loss Landscape Complexity}
\label{sec:landscape_complex}

\begin{figure}[ht]
\centering
\includegraphics[width=0.8\textwidth]{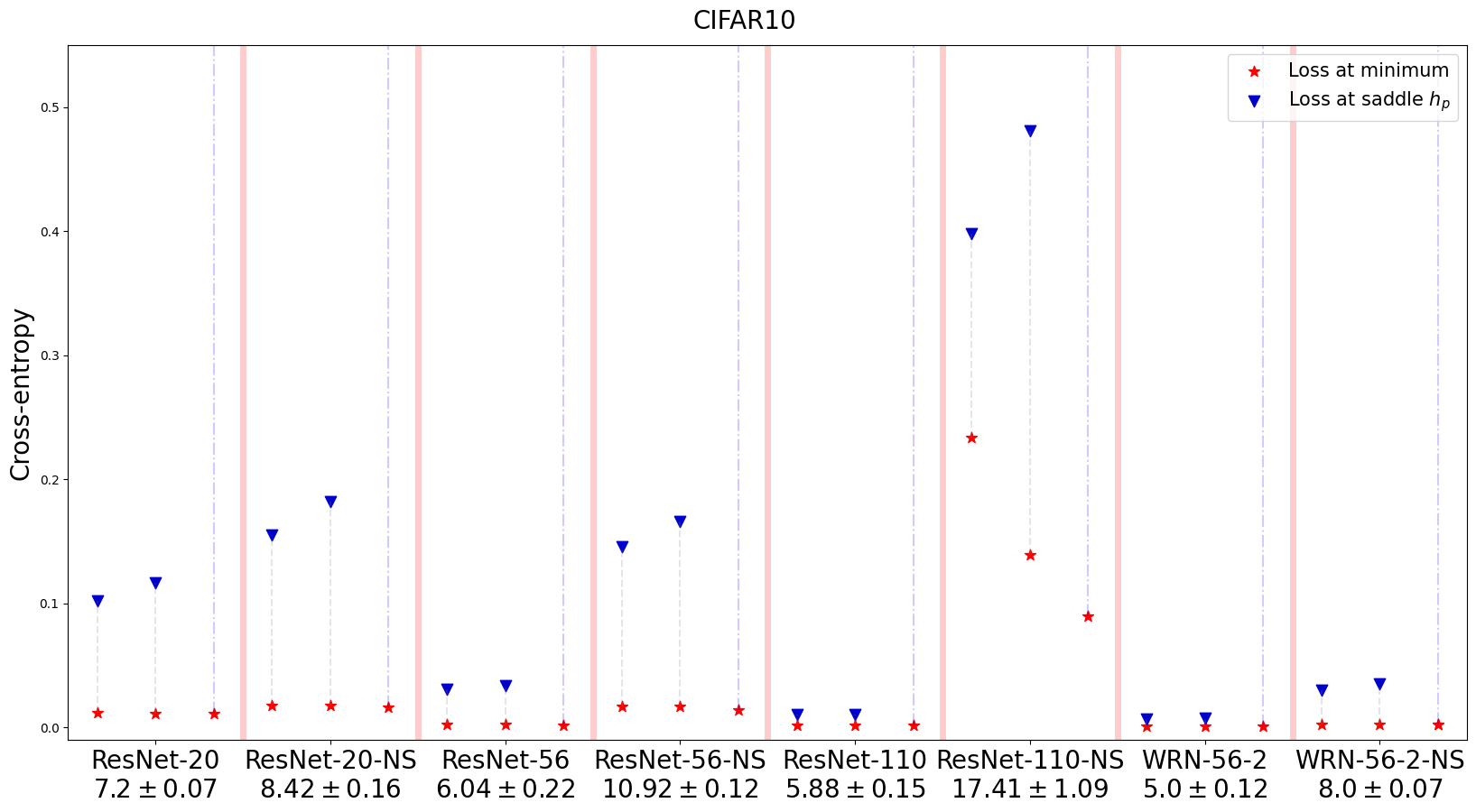}
\caption{Loss barcodes for ResNet-like and VGG-like networks of different depth and width. Model specification is accompanied with average test error.}
\vspace{-.1in}
\label{fig:resnets_barcode}
\end{figure}

This section explores the relationship between the loss barcode and the complexity of loss landscape geometry. In a setup similar to \cite{li2018visualizing}, we investigate whether the loss barcode is capable of capturing the landscape transition from smooth to chaotic behavior. The work \cite{li2018visualizing} empirically studies the effect of depth and width on the loss surface in the vicinity of a local minimum for neural networks with skip-connections (referred to as ``ResNet-like networks'') and without (in \cite{li2018visualizing}, referred to as ``VGG-like networks''). In particular, when depth increases, the work \cite{li2018visualizing} discovers that the loss landscape of VGG-like networks transitions from nearly convex to chaotic, while the loss landscape of ResNet-like networks has no significant change in convexity. Both types of networks benefit from an increase in width as their loss landscape becomes more smooth and convex.

We consider neural networks trained on the CIFAR10 dataset. ResNet-like networks are: ResNet-$20$, ResNet-$56$, ResNet-$110$ which have $20$, $56$, $110$ layers, respectively. VGG-like networks are: ResNet-$20$-NS, ResNet-$56$-NS, ResNet-$110$-NS produced from ResNet-like networks by removing shortcut connections (suffix ``NS'' stands for ``no skip-connection''). Also, we use WideResNet-$56$-$k$(-NS) with $k=1, 2$ meaning k times more filters per layer. WideResNet-$56$-$1$(-NS) is essentially ResNet-$56$-$1$(-NS). 
Figure \ref{fig:resnets_barcode} provides the computed barcodes, Table \ref{tab:landscape_complex:char_mins} in Appendix \ref{app:exp_details} gives more details on the local minima obtained.

With an increase in depth, there is the loss barcode lowering for ResNet-like networks and, by contrast, the increase of barcodes for VGG-like networks. The deepest ResNet-110-NS has the largest barcodes and the largest test error, meaning that without skip-connections this architecture has a hard to optimize complex loss surface. With an increase in width, there is the loss barcode lowering for both types of neural networks, and the loss surfaces are closer to a convex shape.
Although the work \cite{li2018visualizing} explored the vicinity of a local minimum and found somewhat similar results, the notion of loss barcode measures numerically the connectedness of multiple local minima and thus captures the changes in the loss landscape on a global scale. Hence, the loss barcode can be used to monitor the topological differences in the loss geometry when the architecture varies.

\section{Loss Barcodes for Transformers and Large Datasets}\label{sec:transformers}

In this section, we conducted an experiment with the GPT model \cite{radford2018improving}\footnote{https://github.com/karpathy/ng-video-lecture} and a large text OSCAR dataset \footnote{https://oscar-project.org/}. Our goal was to explore a common problem for this area when the size of the data is usually larger than the number of parameters of the model and inspect the corresponding barcodes. We utilized the model of $10$M training parameters and the dataset of $100$M training objects.

\begin{figure}[ht]
\centering
\begin{subfigure}[t]{.49\textwidth}
    \includegraphics[width=\textwidth]{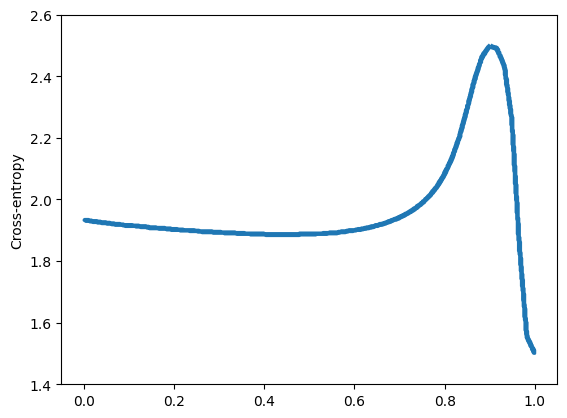}
    \caption{A curve between two different types of minima.}
\end{subfigure}
\begin{subfigure}[t]{.49\textwidth}
    \includegraphics[width=\textwidth]{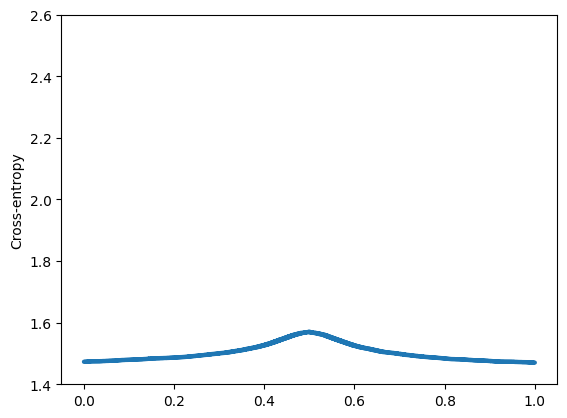}
    \caption{A curve between two identical types of minima.}
\end{subfigure}
\caption{A curve examples between two GPT model`s minima.}
\label{fig:transformers_curves}
\end{figure}

\begin{figure}[ht]
\centering
\includegraphics[width=0.7\textwidth]{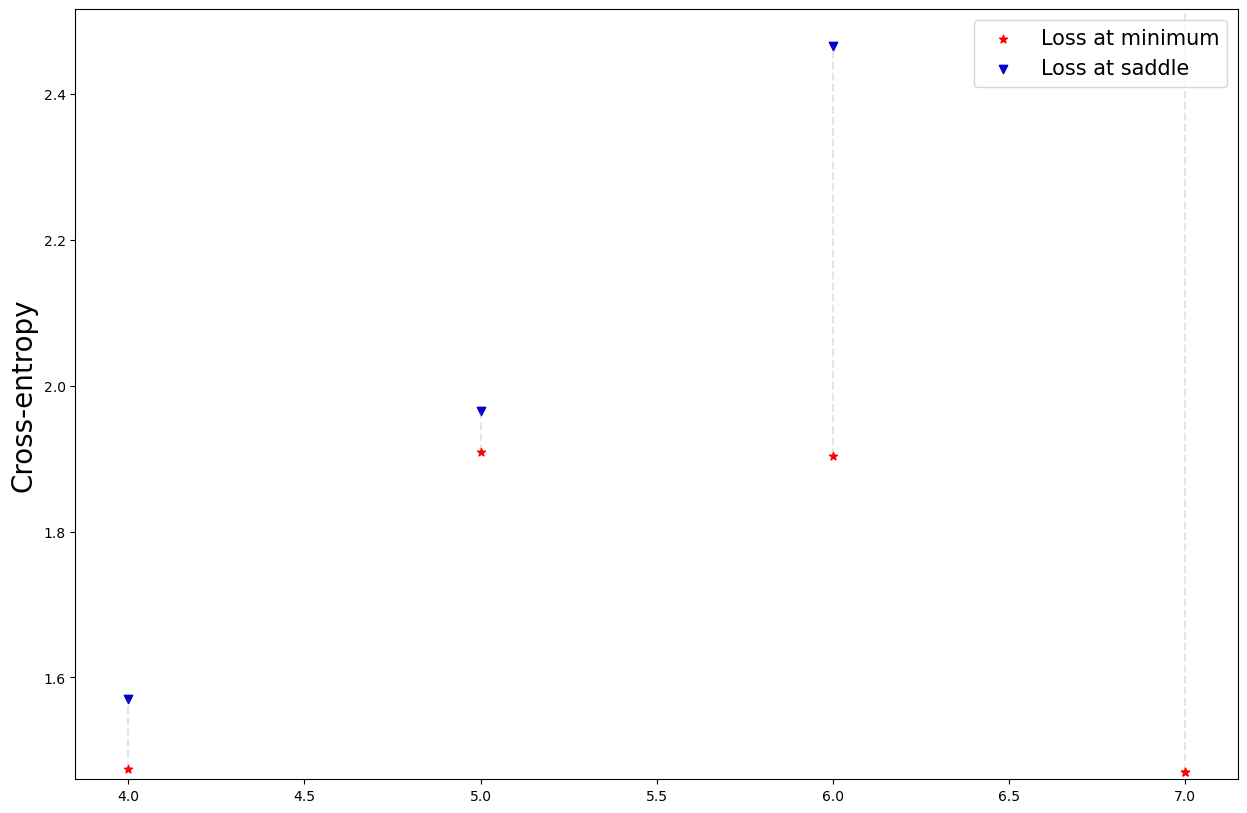}
  \caption{The Loss Barcode of GPT model trained on The OSCAR project dataset for 4 local minima.
  }
  \label{fig:transformer_barcodes}
\end{figure}

We trained the GPT model with four different random seeds. Although we set equal training setup for all random seeds, we obtained four points in parameter space with two distinct levels of quality: high and low. Even with the increased number of epochs, the SGD algorithm was unable to escape from the high minima and find lower ones. According to the Algorithm \ref{algPenlt}, we approximated the paths connecting local minima with the Bezier curves \cite{garipov2018loss} and calculated the training loss along the curves.
As a result, we have two types of curves: between minima with similar loss values and different loss values, see Figure \ref{fig:transformers_curves}. The calculated loss barcodes are shown in Figure \ref{fig:transformer_barcodes} for all local minima together.

Our experiments show that the penalty for higher loss minima to reach the lower minima remains big even after a $5$-times increase in the training budget to find a low loss path, compared to the budget spent searching for local minima.

We deduce that mode connectivity
struggles in the case of transformer architecture. It means that finding a curve between two minima with a low training loss is much more difficult or even impossible for transformer architectures and real datasets of a size larger than the network size. This high-barrier structure is further reflected with computed the loss barcodes.

\section{Scalability and Robustness of Loss Barcodes Calculation}
\label{sec:RandomPointsOptim}

\textbf{Scalability}. In this section, we demonstrate an example of loss barcode computation on the more complex CIFAR100 dataset in a highly overparameterized regime. We used a WideResNet16x10-like architecture \cite{Zagoruyko2016WRN} containing approximately 17.1M of trainable parameters, that is, exceeding the number of training samples more than $300$ times. Our goal is to inspect the behavior of the loss surface under the specified setup of increased complexity and to verify the robustness and scalability of the loss barcode computation procedure.
For this purpose, five WideResNet16x10-like models \cite{Zagoruyko2016WRN} have been trained on the CIFAR100 dataset from different initializations. Table \ref{tab:wrn_char_mins} in the Appendix \ref{app:exp_details} contains the characteristics of the local minima.
The loss barcode is computed according to the Algorithm \ref{algPenlt}.
Detailed training procedures for both minima and path optimization can be found in the Appendix \ref{app:train_hyp}.

\begin{figure}[ht]
\centering
\includegraphics[width=0.7\textwidth]{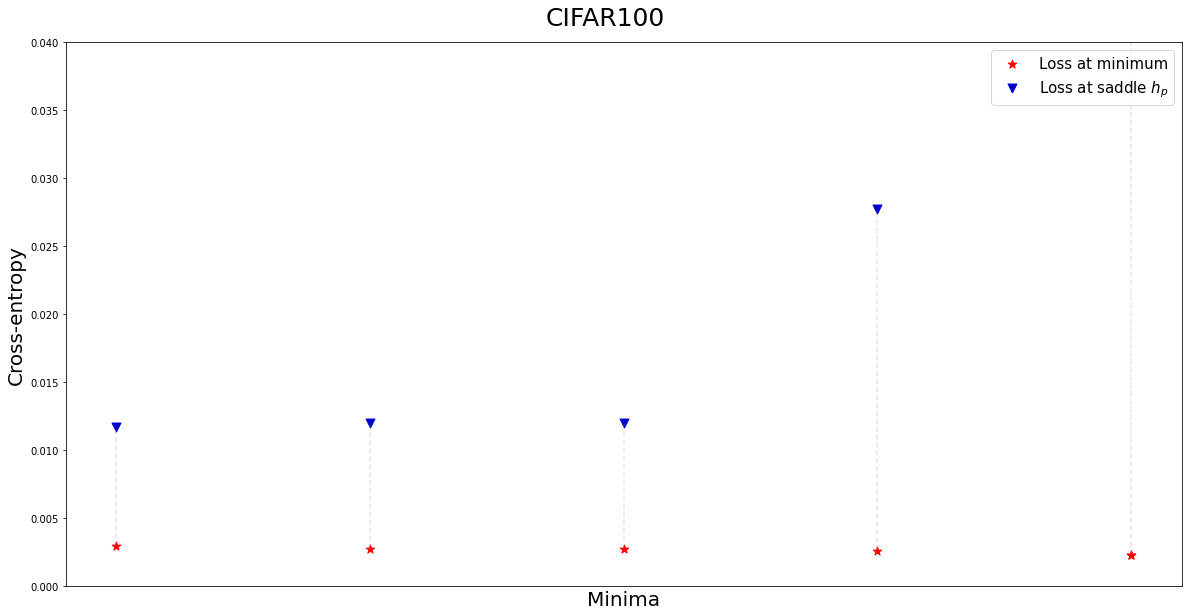}
\caption{Loss Barcodes of WideResNet16x10 trained on the CIFAR100 dataset.}
\label{fig:wrn_barcode}
\end{figure}

The resulting barcode is depicted in Figure \ref{fig:wrn_barcode}.
We highlight that even when the number of trainable parameters is sufficiently large, the loss barcode computation is still feasible. In essence, it is roughly proportional to the training of a single model several times and can be done sequentially or in parallel.

\begin{figure}[!ht]
\centering
\begin{subfigure}[t]{.49\textwidth}
    \includegraphics[width=\textwidth]{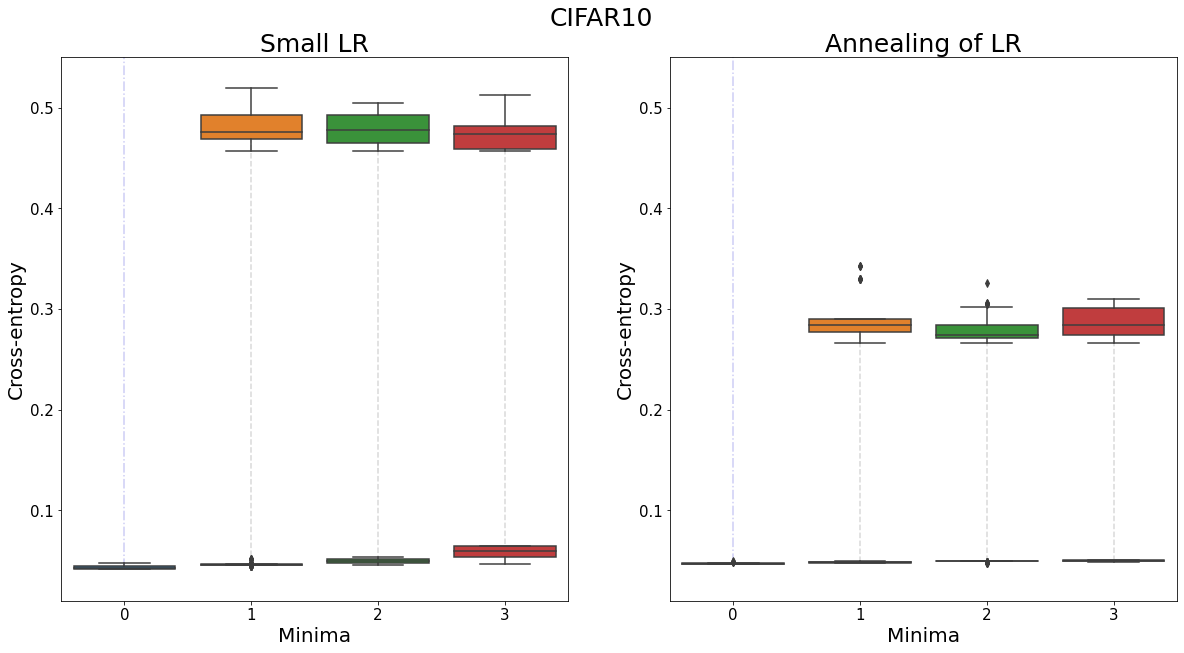}
\end{subfigure}
\begin{subfigure}[t]{.49\textwidth}
    \includegraphics[width=\textwidth]{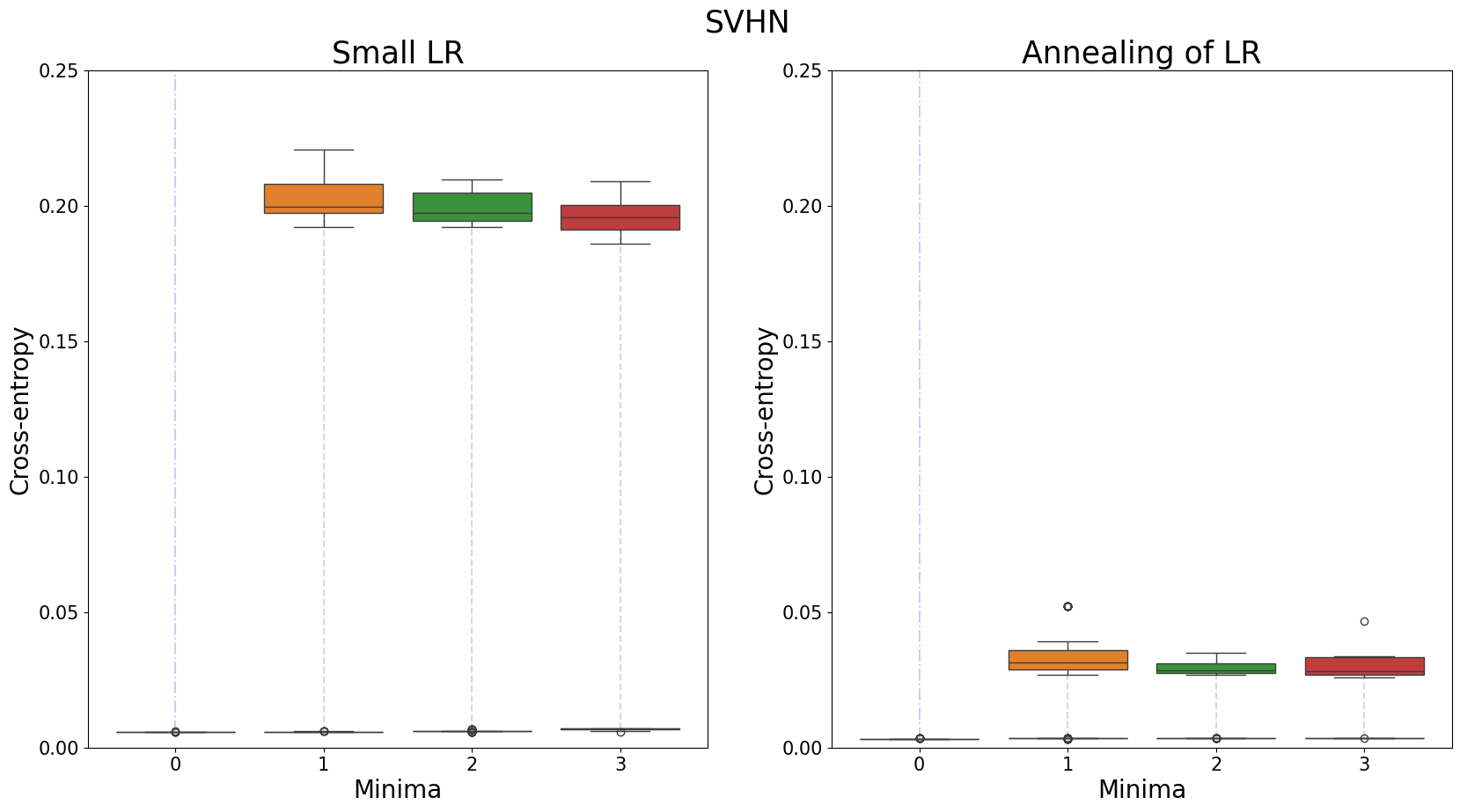}
\end{subfigure}
\caption{Robustness of the barcode computation for ResNet and CIFAR10 (a) and SVHN (b) datasets. For each minimum (x-axis) in the variational series of a sample of size $4$, the lower box shows loss function deviation at the minimum while the upper box shows deviation of the corresponding minimal obligatory penalty.}
\label{fig:resnet9_dev}
\end{figure}

\textbf{Robustness}.
To test the robustness of our algorithm, we conduct experiments using two datasets and the ResNet model; see Figure \ref{fig:resnet9_dev} and Appendix \ref{app:err_bars}. We do this by computing the loss barcodes multiple times for a single architecture and then calculating the confidence intervals. This experiment ensures that the loss barcodes are robust and do not depend on random initializations.

\section{Loss Barcode and Optimal Learning Rate}\label{sec:optim_lr}

The lengths of segments in a loss barcode are known to be equal to the activation energies of the Arrhenius law from the expressions for the exponentially small eigenvalues of the Witten Laplacian \cite{viterbo:2011, peutrec2020barcodespersistentcohomology}. These eigenvalues have recently been shown to play a role in determining the optimal learning rates for stochastic gradient descent \cite{shi2023learning,zhang2017hitting}.
In this direction, we set up a simple experiment and measure the relation between the smallest learning rate, required for escaping the local minimum, and the height of the corresponding segment in the barcode.
The simplified loss landscapes in the form of polynomials are shown in Figure \ref{fig:4polynoms} left (see details in Appendix \ref{app:polynom}). The initial point of the optimization trajectory was fixed in the vicinity of the local minimum. We have found that in this setting, escape learning rates depend essentially linearly on the minimum barcode segment height; see Figure \ref{fig:4polynoms}, right. Interestingly, this is in contrast to the stochastic behavior of the number of steps required for escape; see Figure \nolinebreak \ref{fig:polynom_main_observ} in the Appendix \ref{app:polynom}.

\begin{figure}[ht]
\centering
\includegraphics[width=0.49\textwidth]{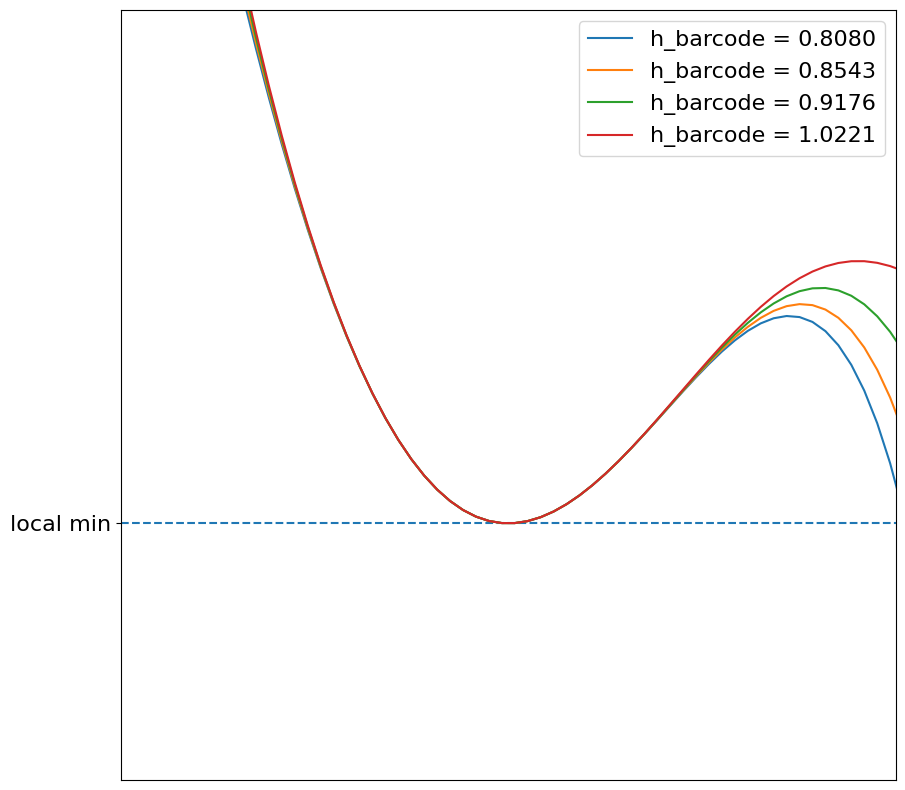}
\includegraphics[width=0.49\textwidth]
{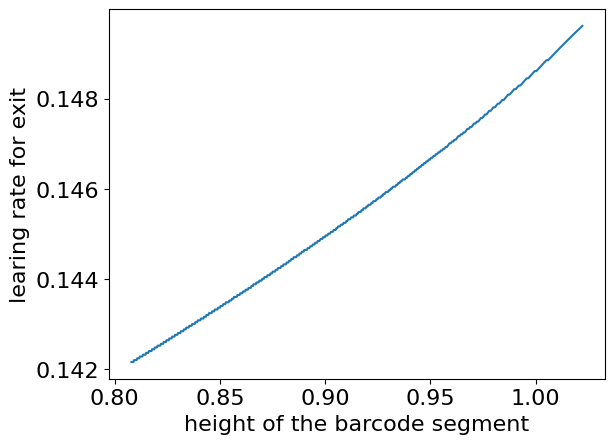}
\caption{For the selected loss landscapes (left), the escape learning rate depends essentially linearly on the the minimum's barcode segment height (right).}
\label{fig:4polynoms}
\end{figure}

\section{Conclusion}
In this work, we have described the topological invariants, loss barcodes, and the neural network's Topological Obstructions (TO-) score that provide the numerical characterization for complexity of loss landscapes in gradient-based learning.
We have shown the pattern of the loss barcodes diminishing and descending lower with increase of the neural network depth and width on various architectures: fully connected, convolutional, transformer, and datasets: MNIST, FMNIST, CIFAR10, CIFAR100, OSCAR. We have observed a connection between the lengths of segments in the loss barcodes and the minima generalization ability. Moreover, the loss barcode captures the increase in the loss landscape complexity when removing skip-connections and is able to reflect topological differences in the loss surface when the underlying architecture is varying. We studied the loss surface of the transformer architecture trained on text data. Its loss surface exhibits a complex nature, having two types of minima without low-loss paths between them.

We consider that the loss barcode will become a helpful tool for analyzing neural network loss surfaces, training dynamics, and generalization ability. We hope that the usage of the loss barcodes will lead to the construction of better neural architectures and training procedures, potentially leading to improved performance.
Also, we envisage numerous scenarios where parameterization-invariant descriptors of loss surfaces might find an application: model selection, designing adversarially robust networks, improving transfer learning, and domain adaptation.

\printbibliography[title={References}]

\appendix

\section{Morse complex and index k critical points' ``lifespans''}\label{morse}
Principal notions of Morse theory are critical points and gradient flow trajectories. The two main objects of gradient-based optimization algorithms are essentially parallel: minima and gradient descent trajectories.  

Here, for the sake of geometric intuitivity, we describe the topological invariants considered under the assumption that the loss is a generic smooth function. The construction can be adopted for the case of continuous piece-wise smooth function or to the discrete case of a function defined on a graph.

\begin{defi}
A point $p$ is a critical point if the gradient at $p$ is zero: $df\vert_p=0$. A critical point $p$ is not degenerate if the Hessian matrix of $f$ at $p$ is non-degenerate.
The index of a non-degenerate critical point is the dimension of the Hessian negative subspace. 
\end{defi}

A gradient flow trajectory $x(t)$, $t\in [t_0,+\infty)$  is a solution to \begin{equation}
   \dot x=-\nabla L(x) \label{eq:gradtraj}  
\end{equation}

A nondegenerate critical point of the index $k$ is  also called $k-$saddle. Near such a point, the function can be written in some coordinates as 
\begin{equation}
   L=-x_1^2\ldots-x_k^2+x_{k+1}^2\ldots+x_n^2.
\end{equation}

The saddles of small index play an important role in the gradient descent optimization algorithms. Gradient descent trajectories are attracted by such points and, more generally, by critical manifolds of a small index. Gradient trajectories spend long periods of time in the vicinities of such points where the gradient is small, while they are trying to find one of the few descending, i.e. negative for the Hessian, directions.

Gradient flow trajectories can be extended back either to converge in the limit $t\to -\infty$ to a critical point or to start from a point at the boundary of the parameter domain. Here and further in this section we assume for simplicity that the gradient of $f$  at each point of the boundary of the domain $\Theta$ is pointing outside the domain, modifications for the general case are straightforward. For each point with non-zero gradient, there is a unique gradient flow trajectory passing through this point. This gives a decomposition of the parameter domain.

The set of all points on the gradient flow trajectories that converge to a given critical point in the limit $t\to -\infty$,  is diffeomorphic to a ball of dimension $k$, where $k$ is the index of the critical point. In fact, first note that local near the given critical point of index $k$ the set of such points lying on the gradient flow trajectories emanating from this critical point is diffeomorphic to a ball of dimension $k$. Next, this diffeomorphism can be extended to the set of all points lying on gradient flow trajectories emanating from this critical point, using the gradient flow. Similarly, the union of all the trajectories emanating from the boundary of the parameter domain is diffeomorphic to $S^{N-1}\times \mathbb{R}$.

Therefore, the parameter domain decomposes into pieces with uniform flow. For each critical point, there is the cell formed by all gradient flow trajectories that converge in the limit $t\to -\infty$ to this critical point. 

Morse complex records how these cells are glued to each other to form the non-trivial part of the decomposition of the total parameter domain.

Let 

\begin{equation}
    \mathcal{M}(p_{\alpha},p_{\beta})=\bigg\{\gamma:\mathbb{R}\rightarrow M^n\bigg\vert\bigg.\
\bigg.\dot{\gamma}=-(\textrm{grad}L)(\gamma(t)),\lim_{t\rightarrow-\infty}=p_{\alpha},\lim_{t\rightarrow+\infty}=p_{\beta}\bigg\} /\mathbb{R}
\end{equation}

be the set of gradient trajectories connecting critical points $p_\alpha$ and $p_\beta$, where the natural action of $\mathbb{R}$ is by the shift $\gamma(t)\mapsto\gamma(t+\tau)$.

If $\textrm{index}(p_{\beta})= \textrm{index}(p_{\alpha})-1$, then generically the set $\mathcal{M}(p_{\alpha},p_{\beta})$ is finite. Let $\#\mathcal{M}(\left[p_{\alpha},\texttt{or}\right],\left[p_{\beta},\texttt{or}\right])$ denote in this case the number of trajectories, counted with signs taking into account a choice of orientation, between critical points $p_\alpha$ and $p_\beta$. Here we have fixed a choice of orientation ``$\texttt{or}$'' of a negative Hessian subspace at each critical point.

Let $\partial_{j}$ denote the matrix with entries $\#\mathcal{M}(\left[p_{\alpha},\textrm{or}\right],\left[p_{\beta},\textrm{or}\right])$.
Each critical point with a choice of orientation can be recorded as a vector (a generator) in vector space.
Then $\partial_{j}$ is the linear operator \begin{equation*}
\partial_{j}\left[p_{\alpha},\texttt{or}\right]=\sum_{\textrm{index}(p_{\beta})=j-1}\left[p_{\beta},\texttt{or}\right]\#\mathcal{M}(p_{\alpha},p_{\beta})
\end{equation*} 

An important property of matrices $\partial_{j}$ is that their consecutive composition is zero:
\begin{equation}
\partial_{j}\circ \partial_{j-1}=0 \label{d2}
\end{equation} 

This can be proved by looking at families of gradient flow trajectories connecting critical points with indexes $\textrm{index}(p_{\beta})= \textrm{index}(p_{\alpha})-2$. Generically, such a family is one-dimensional and compact, and therefore it is a disjoint union of circles and segments. The property \ref{d2} then follows from the fact that each segment has two boundary points, so the corresponding composite gradient trajectories naturally cancel each other out. 

Such sets of matrices $\partial_{j}$ with the property (\ref{d2})  define a chain complex.
Recall that a chain complex $(C_{*},\partial_{*})$ is a sequence
of finite-dimensional vector spaces $C_j$ (spaces of ``$j-$chains'') and linear operators (``differentials'')
\[
\rightarrow C_{j+1}\stackrel{\partial_{j+1}}{\rightarrow}C_{j}\stackrel{\partial_{j}}{\rightarrow}C_{j-1}\rightarrow\ldots\rightarrow C_{0},
\]
which satisfy (\ref{d2}). 
The vector spaces in our case are $C_j=\mathbb{F}^{m_j}$ where $m_j$ is the number of index $j$ critical points and $\mathbb{F}$ is the coefficient field which is most often taken in applications to be $\mathbb{F}=\{0,1\}$ or $\mathbb{Q}$. 
The chain complex can be thought of as the algebraic counterpart of the intuitive idea of representing a complicated geometric object as something decomposed into simple pieces.

The $j-$th homology of the chain complex $(C_{*},\partial_{*})$
is the quotient of vector spaces
\begin{equation}
  H_{j}(C_{*},\partial_{*})=\ker\left(\partial_{j}\right)/\textrm{im}\left(\partial_{j+1}\right).  
\end{equation}
Elements of $\ker\left(\partial_{j}\right)$ are called ``cycles''. 

For any pair of critical points $p_\alpha$,  $p_\beta$ with  $\mathcal{M}(p_{\alpha},p_{\beta})\neq \varnothing$,  we have $L(p_\alpha)> L(p_\beta)$ since the value of the function is always decreasing along the gradient trajectory (\ref{eq:gradtraj}).  
Let $F_rC_j\subseteq C_j$ denote the subspace of $C_j$ spanned by the critical points of the index $j$ with the critical value $L(p_\alpha)\leq r$. It follows that: 
\begin{equation}\partial_j(F_rC_j)\subset F_rC_{j-1}.\label{eq:df}\end{equation} These subspaces $F_rC_j$ are nested: $F_{r_1}C_{j}\subseteq F_{r_2}C_{j}$ for any pair $r_1< r_2$. 

The homology of the chain complex $(F_rC_{*},\partial_{*})$ is the homology of the subset $\Theta_{L\leq r}=\{L(x)\leq r\vert x\in\Theta\}$ of the parameter domain. In particular, the dimension of the $j-$th homology $H_j(F_rC_{*},\partial_{*})$ is equal to the number of independent $j-$dimensional topological features in $\Theta_{L\leq r}$. 

As $r$ increases, the dimension of homology $H_*(F_rC_{*},\partial_{*})$ can increase or decrease. These changes can only occur at $r$ equal to one of the critical values of the loss function.  The ``Main'' theorem on persistent homology, also sometimes called ``Structure'' or ``Principal'' theorem on persistent homology, \cite{B94,barannikov2021canonical,zomorodian2001computing,viterbo:2011} describes these changes in terms of simple ``birth''-``death'' phenomena of cycles chosen uniformly across all values of $r$.  

The relevant notion to formulate this result is the notion of filtration on the chain complex. The set of nested subspaces $(F_{r_\alpha}C_*)$ indexed by a subset $\{r_\alpha\}\subset \mathbb{R}$ and satisfying (\ref{eq:df}) is called an ``$\mathbb{R}-$filtration''. Then the mentioned theorem can be described as the classification theorem for $\mathbb{R}-$filtered chain complexes.  

It follows that the critical values of the loss function, except for the global minimum, are canonically split into pairs. We note that this fact generalizes straightforwardly also to the case of critical manifolds and/or piece-wise smooth loss functions.

This splitting implies in particular that each $k-$saddle of the loss function either kills a topological feature that was born at a lower $(k-1)-$saddle or gives birth to a topological feature that is killed at a higher $(k+1)-$saddle. In both cases, this gives the ``lifespan'' bars of the barcode associated with these pairs of critical points. We expect that for critical points of small indexes these ``lifespan'' bars from the higher barcodes are also located in a small lower part of the range of the loss values and the ``lifespans'' of critical points of small indexes diminish with increasing depth and width of deep neural networks.

\section{TO-score and convexity up to reparameterization}
\label{app:to_score}

\textbf{Theorem \ref{th:th1}}
Let $L$ be a piece-wise smooth continuous function on a domain $D\subset \mathbb{R}^n$, $n\geq 5$, with $-\nabla(L)\vert_{\partial D}$ pointing outside the domain $D$, and such that for all $r\geq 0$ index $r$ TO-score$(L)=0$.  Then there exists an arbitrary small smooth perturbation of $L$ which is convex after a smooth reparameterization of the domain $D$.
\begin{proof}
  First, there exists an arbitrary small smooth perturbation $\tilde{L}$ of $L$ whose all critical points are non-degenerate by Stone-Weierstrass and Sard's theorems. Then,  because of stability of the barcode, see e.g. \cite{chazal2021introduction}, the vanishing of all TO-scores of $L$ implies that all $\tilde{L}$ critical points, except for the global minimum, come in pairs with small ``lifespans''. 
  Then the first, second, and third elimination theorems from \cite{milnor2015lectures}, see also \cite{smale},
  allow to eliminate all these pairs of critical points and the constructed perturbation is small because of the smallness of the critical ponts lifespans. This gives a smooth arbitrary small perturbation $\hat{L}$ of $L$ whose only critical point is the non-degenerate global minimum. Then a diffeomorphism that makes $\hat{L}$ convex is constructed with the help of the $\hat{L}$ gradient flow. That is, the $\hat{L}$ gradient flow can be used to bring the domain $D$ to a neighborhood of the critical point, while rescaling the function, and by the Morse lemma $\hat{L}$ is quadratic at some coordinates in a neighborhood of its unique critical point.
\end{proof}

\section{Visual Interpretation of the Insertion Procedure} \label{app:insert}
\begin{figure}
\centering
  \includegraphics[width=0.6\textwidth]{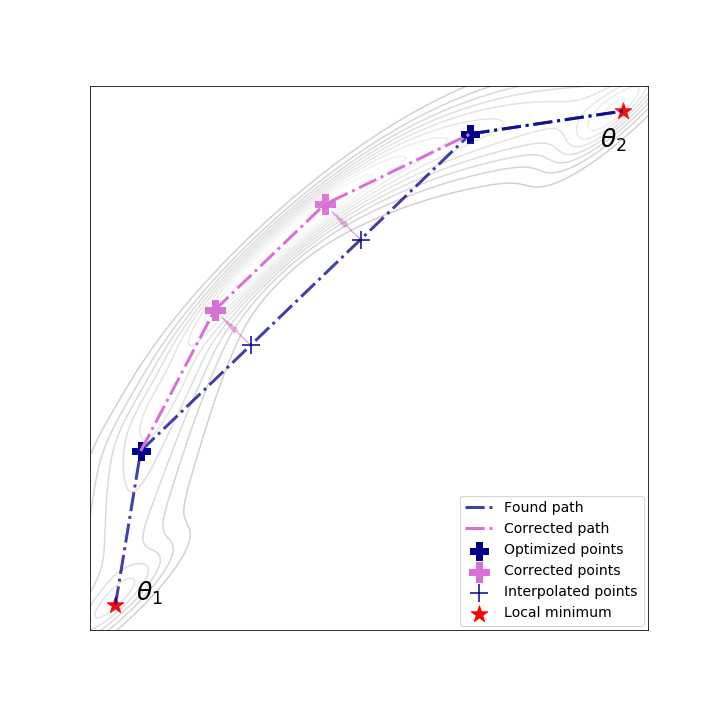}
  \vskip-30pt
  \caption{Illustration of point insertion.}
  \label{fig:PI_model}
  \vskip-5pt
\end{figure}
We motivate the insertion procedure proposed in Section \ref{gradflowsegm} with the following example on the model data. In the first stage, two points were optimized, and they ended in reasonable regions of loss function with low loss values. However, it is visually noticeable that linear interpolation between them sometimes produces points completely off the desired manifold, preserving low values of loss function. However, applying our procedure, by adding new points and optimizing them, they will reach the relevant region and we can eliminate such a negative effect, which can be seen in Figure \ref{fig:PI_model}.

\section{Barcodes and the Exit Learning Rate Experiment}\label{app:polynom}

In this section, we conducted an experiment with polynomial function in order to see the dependency between the height of the barcode's segments and the optimal learning rate for escaping the local minima of the loss landscape. We consider a polynomial of degree six as an example of loss landscape and use two approaches to adjust the height of the barcode's segment for a local minimum of this polynomial, see Figure \ref{fig:polynom_function} and Figure \ref{fig:4polynoms} (left).

We randomly sampled a point in the vicinity of a local minimum and performed gradient descent optimization. We tracked the optimization trajectory and the moment of escape from the local minimum (jumping over the saddle point), see Figure \ref{fig:polynom_trajectories}.

\begin{figure}[ht]
\centering
\begin{subfigure}[t]{.7\textwidth}
    \includegraphics[width=\textwidth]{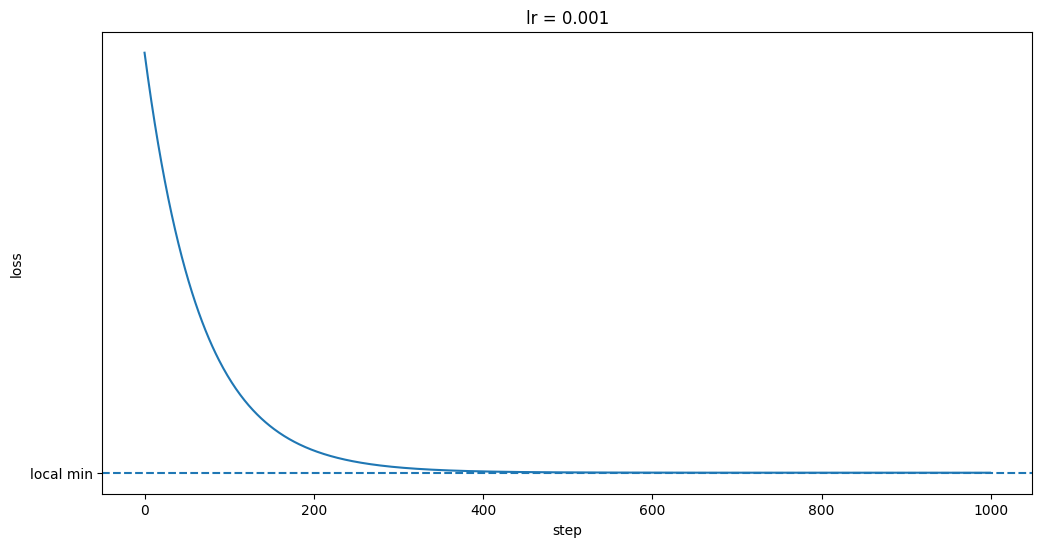}
    \caption{GD converges to local minimum.}
\end{subfigure}
\begin{subfigure}[t]{.7\textwidth}
    \includegraphics[width=\textwidth]{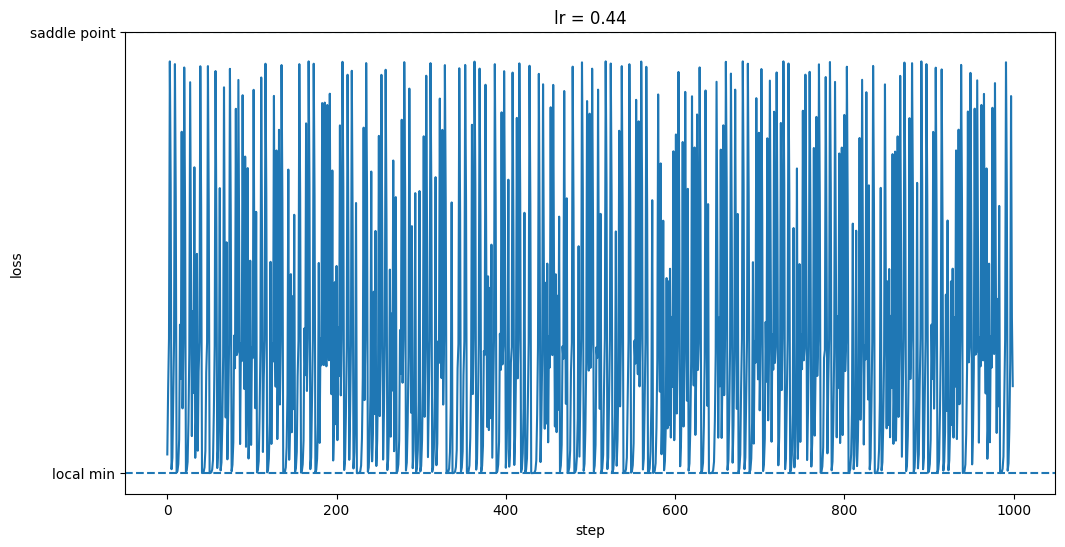}
    \caption{GD cannot converge to local minimum and cannot escape it.}
\end{subfigure}
\begin{subfigure}[t]{.7\textwidth}
    \includegraphics[width=\textwidth]{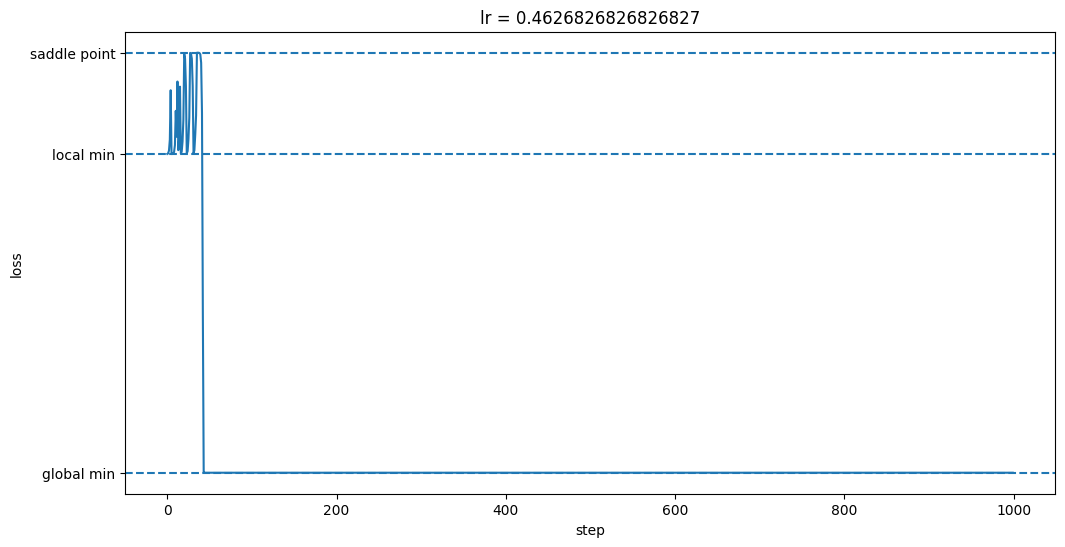}
    \caption{GD quickly escapes local minimum and converges to global minimum.}
\end{subfigure}
\caption{ Types of behavior of GD near a local minimum.}
\label{fig:polynom_behaviour}
\end{figure}

\begin{figure}[ht]
\centering
\begin{subfigure}[t]{0.8\textwidth}
    \includegraphics[width=\textwidth]{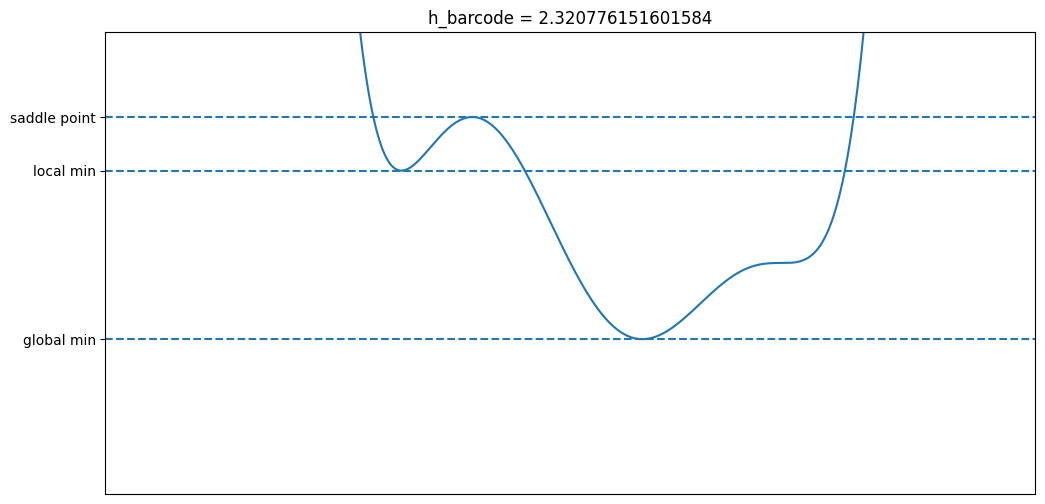}
    \caption{Sixth degree fixed polynomial.}
\end{subfigure}
\caption{Second type of considered loss landscapes.}
\label{fig:polynom_function}
\end{figure}

\begin{figure}[ht]
\centering
\begin{subfigure}[t]{0.49\textwidth}
    \includegraphics[width=\textwidth]{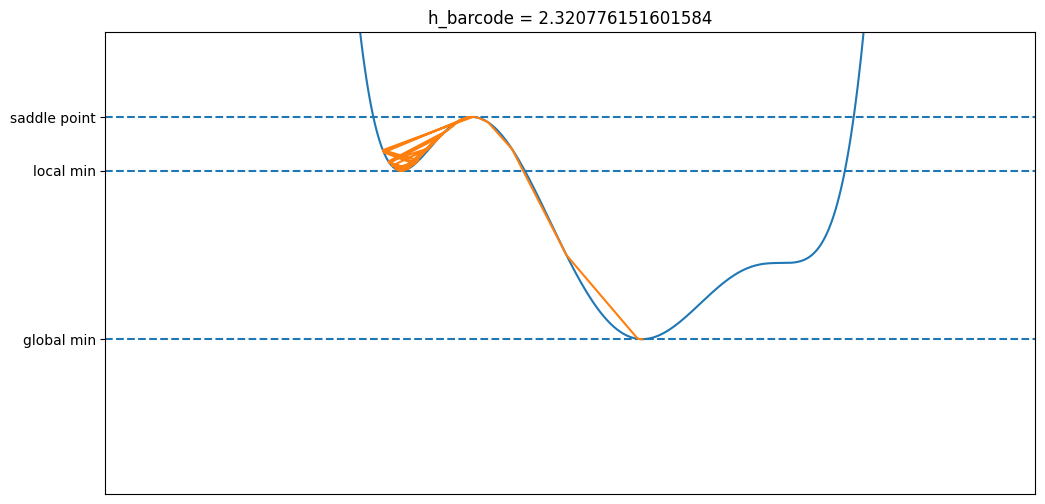}
    \caption{First approach with two points.}
\end{subfigure}
\begin{subfigure}[t]{0.49\textwidth}
    \includegraphics[width=\textwidth]{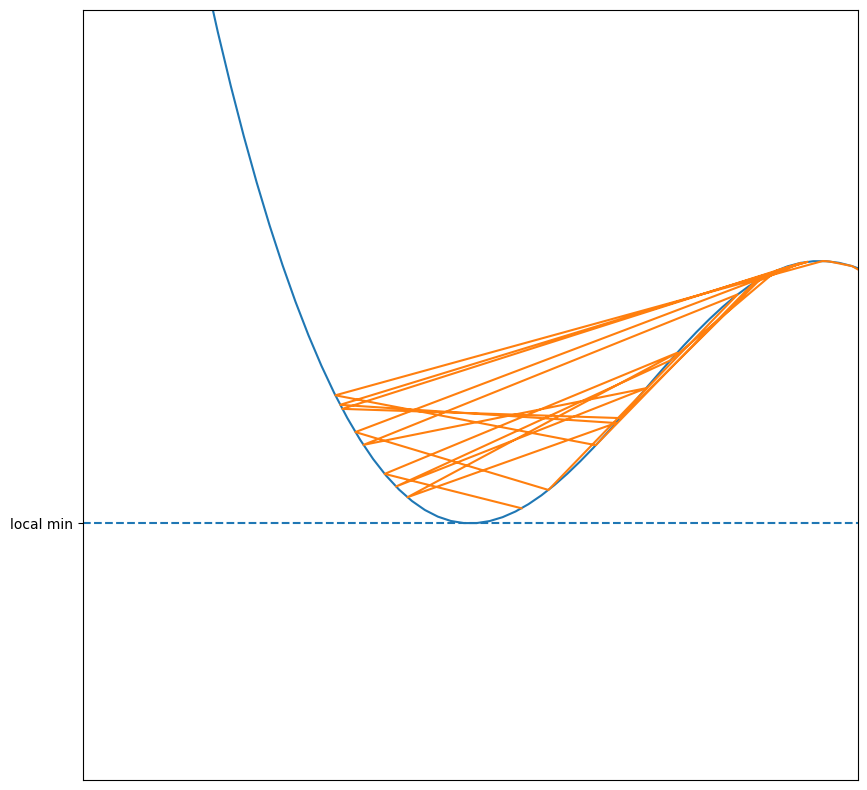}
    \caption{Second approach with only one controllable point.}
\end{subfigure}
\caption{Sixth degree polynomial and the trajectories of GD which are able to escape local minimum (two approaches).}
\label{fig:polynom_trajectories}
\end{figure}

In the first approach, we have values of up to the second derivative at the two points of the local minimum and the global minimum. These experiments revealed that there are three regimes of the gradient descent algorithm to leave the local minimum depending on the learning rate value, see Figure \ref{fig:polynom_behaviour}.

\begin{figure}[ht]
\centering
{\includegraphics[width=0.6\linewidth]{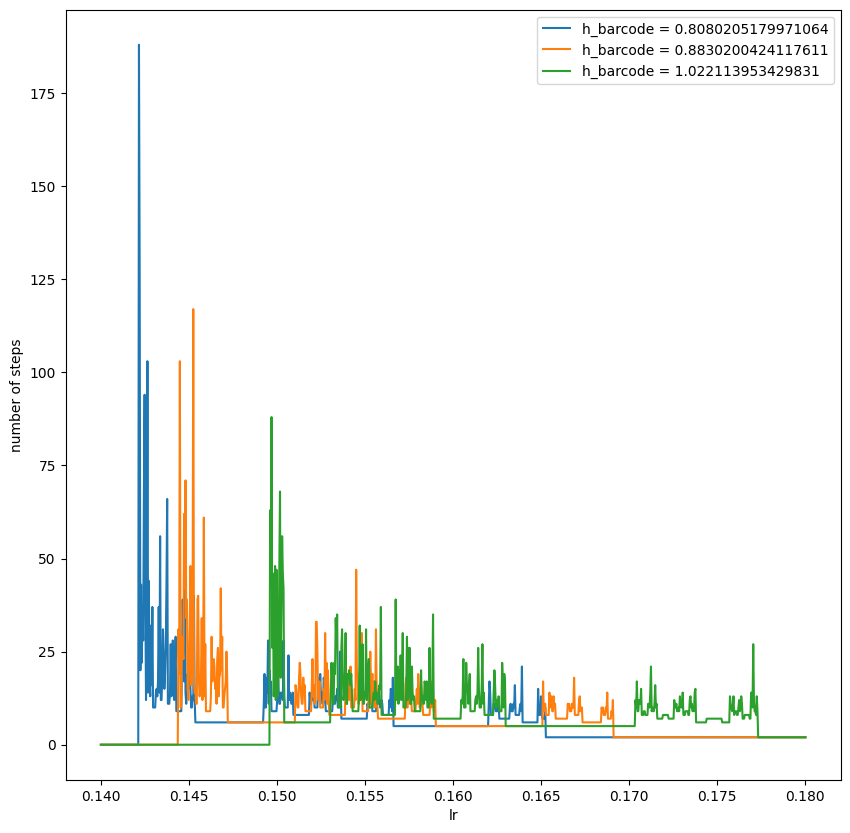}}
\caption{The relationship between the number of steps to exit the local minimum and the learning rate (with different heights of barcode segment, zero number of steps means that GD cannot get out).}
\label{fig:polynom_main_observ}

\end{figure}

In the second approach, all values of the polynomial derivatives up to the fifth order at the given local minimum were fixed. This allowed us to accurately vary the height of the barcode without impacting the landscape of the loss function near the local minimum. The dependence of the height of the barcode segment on the learning rate necessary to escape the local minimum was investigated; see Figure \ref{fig:polynom_main_observ}. In addition, for a fixed height of the barcode, we explore the behavior of gradient descent with respect to the learning rate value.

\section{Error Bars}\label{app:err_bars}
\begin{figure}[!ht]
\centering
  \includegraphics[width=0.8\textwidth]{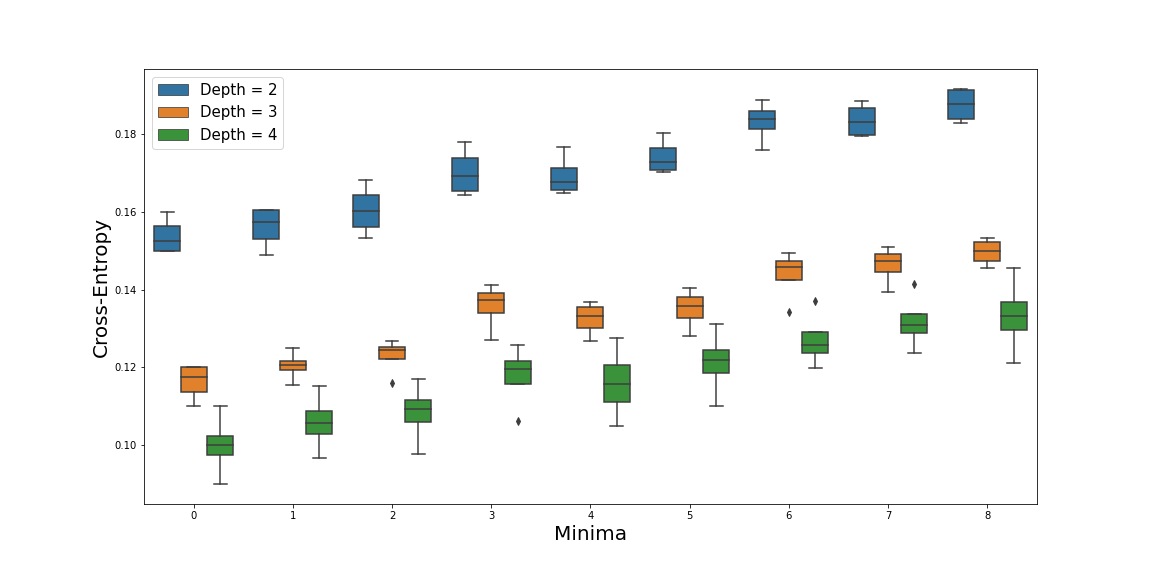}
  \caption{Error bars in the experiment with FC deep neural networks with 2,3, and 4 hidden layers. The minima are ordered in the same order as on Figure \ref{fig:network-barcodesMnist}}
  \label{fig:ErrB}
\end{figure}

We report the error bars in the experiments with FC deep neural networks in Figure \ref{fig:ErrB}. Standard deviations are calculated for the values of the maxima of loss on the optimized curves that connect the minima.

\section{Training Details}
\label{app:train_hyp}
This section provides training details for the experiments presented in this work.

\paragraph{Barcodes and Minima's Generalization Ability, Section \ref{sec:exp_gener_ability}.} In this paragraph, we describe the training procedures that have been used for experiments with ResNet-like architecture and the CIFAR10, SVHN datasets from Section \ref{sec:exp_gener_ability}. The learning rate schedule is given by the following formula:
\begin{equation*} \label{eq1}
\begin{split}
\begin{cases} lr^{max}, & \mbox{if } b_{i} \le |D| m_1\\
 (1 - \Delta_i) \cdot lr^{max} + \Delta_i \cdot lr^{min}, & \mbox{if } |D| m_1 < b_{i} < |D| m_2  \\
 lr^{min}, & \mbox{if } b_{i} \ge |D| m_2\end{cases}
\end{split}
\end{equation*}
where $|D|$ -- number of batches in the dataset and $\Delta_i = \frac{b_{i} - |D| m_1}{|D|(m_2 - m_1)}$. Without loss of generality, this system of conditions can be expressed as a function $S(m_1, m_2, lr^{max}, lr^{min}, i)$. Note that in such a formulation, $S(0, 0, lr_0, lr_0, \cdot)$ corresponds to the simplest case when the learning rate is fixed to the value $lr_0$ throughout the training process. We denote it by $S(lr_0)$ for simplicity. We have used schedulers $S_{c10}:=S(10, 80, 10^{-2}, 10^{-4}, \cdot)$ and $S_{svhn}:=S(50, 100, 5 \cdot 10^{-2}, 10^{-4}, \cdot)$ to optimize local minima on the CIFAR10 and SVHN datasets (Table \ref{tab:exp_gener_ability}), respectively, in combination with Stochastic Gradient Descent and momentum $\nu=0.9$. We have used a path optimization procedure in the form of a polygonal chain with one bend from \cite{garipov2018loss} with a training procedure of type $2$ from Table \ref{tab:exp_gener_ability}. For the experiments, we used a single Nvidia Titan RTX GPU.

\begin{table}
  \caption{Training hyperparameters used for CIFAR10 and SVHN experiments in Section \ref{sec:exp_gener_ability}.}
  \label{tab:exp_gener_ability}
  \centering
  \begin{tabular}{lllll}
    \toprule
    Type & Batch size & $\|\theta\|_2$ & Scheduler & Epochs  \\
    \midrule
    \multicolumn{5}{c}{CIFAR10} \\
    \midrule
    $1$ & $256$ & $5 \times 10^{-3}$ & $S(10^{-4})$ & $4500$     \\
    $2$ & $256$ & $5 \times 10^{-3}$ & $S_{c10}$ & $300$      \\
    \midrule 
    \multicolumn{5}{c}{SVHN} \\
    \midrule
    $1$ & $256$ & $5 \times 10^{-4}$ & $S(10^{-4})$ & $1500$     \\
    $2$ & $256$ & $5 \times 10^{-4}$ & $S_{svhn}$ & $100$     \\    
    \bottomrule
  \end{tabular}
\end{table}

\paragraph{Barcode as a Measure of Loss Landscape Complexity, Section \ref{sec:landscape_complex}.} In experiments with ResNet-like and VGG-like neural networks and the CIFAR10 dataset from Section \ref{sec:landscape_complex}, we have closely followed the experimental setup from \cite{li2018visualizing}. In particular, we have used the training procedures from \cite{li2018visualizing} to optimize a set of local minima. For each of the neural networks, the same training procedure have been used for path optimization in the form of polygonal chain with one bend.

\paragraph{Loss Barcodes in Overparametrized Regime, Section \ref{sec:RandomPointsOptim}.} In the experiment with the WideResNet16x10 and CIFAR100 dataset fin Section \ref{sec:RandomPointsOptim}, we have used the training procedure for WideResNet16x10 and CIFAR100 from the paper \cite{garipov2018loss} to optimize a set of local minima. The same training procedure has been used for path optimization in the form of a polygonal chain with one bend.

\section{Architectures of DNNs}
\label{app:architectures}

Tables \ref{tab:cnn1_c}, \ref{tab:cnn2_c}, \ref{tab:cnn3_c} present the architectures of convolutional neural networks from the experiments in Section \ref{sec:explowering}. The ResNet architecture from the experiments in Section \ref{sec:exp_gener_ability} is depicted in Table \ref{tab:exp_gener_ability:res_arch}. In Section \ref{sec:landscape_complex}, we have used the ResNet-like and VGG-like architectures from Section 6 in \cite{li2018visualizing}.

\begin{table}[ht]
\caption{Architecture of CNN-1-C used in the Section \ref{sec:explowering} experiments.} 
\centering
\begin{tabular}{@{}l@{}}
\toprule
\multicolumn{1}{c}{CNN-1-C} \\ \midrule
$x^0$ = Conv(3, C, 5x5), ReLU \\ \midrule
$x^1$ = MaxPooling(8x8) \\ \midrule
$\hat{y}$ = Linear(16 $\cdot$ C, 10) \\ \bottomrule
\end{tabular}
\label{tab:cnn1_c}
\end{table}

\begin{table}[ht]
\caption{Architecture of CNN-2-C used in the Section \ref{sec:explowering} experiments.} 
\centering
\begin{tabular}{@{}l@{}}
\toprule
\multicolumn{1}{c}{CNN-2-C} \\ \midrule
$x^0$ = Conv(3, C, 5x5), ReLU \\ \midrule
$x^1$ = MaxPooling(2x2) \\ \midrule
$x^2$ = Conv(C, C, 5x5), ReLU \\ \midrule
$x^3$ = MaxPooling(4x4) \\ \midrule
$\hat{y}$ = Linear(16 $\cdot$ C, 10) \\ \bottomrule
\end{tabular}
\label{tab:cnn2_c}
\end{table}

\begin{table}[ht]
\caption{Architecture of CNN-3-32 used in the Section \ref{sec:explowering} experiments.}
\centering
\begin{tabular}{@{}l@{}}
\toprule
\multicolumn{1}{c}{CNN-3-C} \\ \midrule
$x^0$ = Conv(3, 32, 5x5), ReLU \\ \midrule
$x^1$ = MaxPooling(2x2) \\ \midrule
$x^2$ = Conv(32, 32, 5x5), ReLU \\ \midrule
$x^3$ = MaxPooling(2x2) \\ \midrule
$x^4$ = Conv(32, 32, 5x5), ReLU \\ \midrule
$x^5$ = MaxPooling(2x2) \\ \midrule
$\hat{y}$ = Linear(16 $\cdot$ 32, 10) \\ \bottomrule
\end{tabular}
\label{tab:cnn3_c}
\end{table}

\begin{table}[ht]
\caption{ResNet architecture used for CIFAR10 and SVHN experiments in the Section \ref{sec:exp_gener_ability}.} 
\label{tab:exp_gener_ability:res_arch}
\centering
\begin{tabular}{@{}l@{}}
\toprule
\multicolumn{1}{c}{ResNet} \\ \midrule
$x^0$ = Conv(3, 16, 3x3), ReLU \\ \midrule
$x^1$ = Conv(16, 16, 3x3), ReLU, Conv(16, 16, 3x3) + $x^0$ \\ \midrule
$x^2$ = MaxPooling(2x2), Conv(16, 32, 3x3), ReLU \\ \midrule
$x^3$ = Conv(32, 32, 3x3), ReLU, Conv(32, 32, 3x3) + $x^2$ \\ \midrule
$x^4$ = MaxPooling(2x2), Conv(32, 64, 3x3), ReLU \\ \midrule
$x^5$ = Conv(64, 64, 3x3), ReLU, Conv(64, 64, 3x3) + $x^4$ \\ \midrule
$x^6$ = MaxPooling(2x2), Conv(64, 64, 3x3), ReLU \\ \midrule
$x^7$ = Conv(64, 64, 3x3), ReLU, Conv(64, 64, 3x3) + $x^6$ \\ \midrule
$x^8$ = MaxPooling(2x2), BatchNorm(64) \\ \midrule
$\hat{y}$ = Linear($X^8$, 10) \\ \bottomrule
\end{tabular}
\end{table}

\begin{algorithm}[ht]
\caption{Refine path}
\label{algo:insertion}
\begin{algorithmic}
\STATE {\bfseries procedure} INSERTION(path, criterion):
\STATE \hskip1em $\mu \gets  \frac{\|\theta_{last} - \theta_{first}\|_2}{\lvert path \rvert}$
\STATE \hskip1em $L_{max} \gets max_{i}(L(\theta_{i})_{f, D})$ where $i \in [0, \lvert path \rvert - 1]$
\STATE \hskip1em $T \gets 1.2$, $i \gets 0$, $B \gets array()$, $D \gets array()$

\STATE \hskip1em {\bfseries while} {$i < \lvert path \rvert - 1$} {\bfseries do}
     \STATE \hskip2em $D[i] \gets criterion(\theta_{i}, \theta_{i+1})$
     \STATE \hskip2em $i:= i+1$
\STATE \hskip1em  {\bfseries end while}
    
\STATE \hskip1em {\bfseries if} {$max(D) > T$} {\bfseries then}
    \STATE \hskip2em $max_{1} \gets argmax_{i}(D)$
    \STATE \hskip2em $max_{2} \gets argmax_{j}(D \backslash \{ max(D) \})$
    \STATE \hskip2em $path[max_{1}] \gets \frac{\theta_{max_{1}} + \theta_{max_{1} + 1}}{2}$
    \STATE \hskip2em $path[max_{2}] \gets \frac{\theta_{max_{2}} + \theta_{max_{2} + 1}}{2}$
    \STATE \hskip2em \text{Length $|path| \uparrow$ 2}
    \STATE \hskip2em $B \gets path_{0}$
    \STATE \hskip2em $path = path \backslash \{ path_{0} \}$
    \STATE \hskip2em $B \gets path_{\lvert path \rvert}$
    \STATE \hskip2em $path = path \backslash \{path_{\lvert path \rvert} \}$ 
    \STATE \hskip2em \text{Length $|path| \downarrow$ 2}
\STATE \hskip1em {\bfseries end if}
\STATE \hskip1em {\bfseries return} path, B
\STATE {\bfseries end procedure}

\STATE 
\STATE {\bfseries procedure} $\text{LOSS CRITERION}_{L_{max}}$ ($\theta_{i}, \theta_{i+1}$)
    \STATE \hskip1em loss = $L((1 - \alpha) \theta_i + \alpha \theta_{i+1})_{f, D}$, where $\alpha \in (0, 1)$
    \STATE \hskip1em {\bfseries return} $\frac{max(loss)}{L_{min}}$
\STATE {\bfseries end procedure}

\STATE

\STATE {\bfseries procedure} $\text{DISTANCE CRITERION}_{\mu}(\theta_{i}, \theta_{i+1})$
    \STATE \hskip1em {\bfseries return} $\frac{\|\theta_{i} - \theta_{i+1}\|_2}{\mu}$
\STATE {\bfseries end procedure}

\end{algorithmic}
\end{algorithm}

\section{Characteristics of Optimized Local Minima and Paths}
\label{app:exp_details}

Table \ref{tab:exp_gener_ability:char_mins} presents the minima characteristics for the CIFAR10 and SVHN experiments from Section \ref{sec:exp_gener_ability}.
Table \ref{tab:landscape_complex:char_mins} presents the minima characteristics for ResNet-like and VGG-like networks in the experiments of Section \ref{sec:landscape_complex}.
Table \ref{tab:wrn_char_mins} presents the minima characteristics for WideResNet16x10 trained on the CIFAR100 dataset.

\begin{table}
  \caption{WideResNet16x10 CIFAR100 minima}
  \label{tab:wrn_char_mins}
  \centering
  \begin{tabular}{llll}
    \toprule
    Train loss & Train acc, \% & Test loss & Test acc, \% \\
    \midrule
    $0.00276 \pm 0.00053$ & $99.98080 \pm 0.00349$ & $0.90929 \pm 0.01299$ & $77.92800 \pm 0.26649$ \\
    \bottomrule
  \end{tabular}
\end{table}
  
\begin{table*}
  \caption{Minima characteristics for CIFAR10 and SVHN experiments from the Section \ref{sec:exp_gener_ability}.}
  \label{tab:exp_gener_ability:char_mins}
  \centering
  \begin{tabular}{lllll}
    \toprule
    Type & Train loss & Train acc, \% & Test loss & Test acc, \% \\
    \midrule
    \multicolumn{1}{c}{} & \multicolumn{4}{c}{CIFAR10}\\
    \midrule 
    Type $1$ & $0.04926 \pm 0.00687$ & $99.95300 \pm 0.03544$ & $1.01237 \pm 0.03886$ & $72.66125 \pm 0.82658$ \\
    Type $2$ & $0.04865 \pm 0.00120$ & $99.98925 \pm 0.00386$ & $0.63170 \pm 0.00770$ & $80.83000 \pm 0.25070$ \\
    \midrule
    \multicolumn{1}{c}{} & \multicolumn{4}{c}{SVHN}\\
    \midrule
    Type $1$ & $0.00611 \pm 0.00052$ & $99.99659 \pm 0.00236$ & $0.42377 \pm 0.01645$ & $89.99789 \pm 0.37510$ \\
    Type $2$ & $0.00341 \pm 0.00019$ & $99.99744 \pm 0.00045$ & $0.23817 \pm 0.01015$ & $94.22057 \pm 0.17428$ \\ 
    \bottomrule
  \end{tabular}
\end{table*}

\setlength{\tabcolsep}{3pt}
\begin{table*}
  \caption{Minima characteristics for ResNet-like and VGG-like networks in the experiments from the Section \ref{sec:landscape_complex}.}
  \label{tab:landscape_complex:char_mins}
  \centering
  \begin{tabular}{lllll}
    \toprule
    Model & Train loss & Train acc, \% & Test loss & Test acc, \% \\
    \midrule
     ResNet-20 & $0.01139 \pm 0.00014$ & $99.89133 \pm 0.00340$ & $0.26323 \pm 0.00315$ & $92.80000 \pm 0.07071$ \\
     ResNet-20-NS & $0.01694 \pm 0.00077$ & $99.69533 \pm 0.01914$ & $0.32643 \pm 0.00856$ & $91.58667 \pm 0.17016$ \\
     ResNet-56 & $0.00211 \pm 0.00009$ & $99.99067 \pm 0.00525$ & $0.25668 \pm 0.00805$ & $93.96667 \pm 0.21746$ \\
     ResNet-56-NS & $0.01638 \pm 0.00109$ & $99.54000 \pm 0.02903$ & $0.48965 \pm 0.00660$ & $89.08667 \pm 0.13123$ \\
     ResNet-110 & $0.00128 \pm 0.00002$ & $99.99533 \pm 0.00094$ & $0.26657 \pm 0.00586$ & $94.12333 \pm 0.15173$ \\
     ResNet-110-NS & $0.15296 \pm 0.05768$ & $94.91733 \pm 1.89809$ & $0.65718 \pm 0.08450$ & $82.60000 \pm 1.09044$ \\
     WideResNet-56-2 & $0.00099 \pm 0.00017$ & $99.99867 \pm 0.00094$ & $0.21582 \pm 0.01029$ & $95.00000 \pm 0.12028$ \\
     WideResNet-56-2-NS & $0.00190 \pm 0.00010$ & $99.98800 \pm 0.00283$ & $0.40339 \pm 0.00455$ & $91.99667 \pm 0.06600$ \\
    \bottomrule
  \end{tabular}
\end{table*}

\begin{figure}
\centering
  \includegraphics[width=0.8\linewidth]{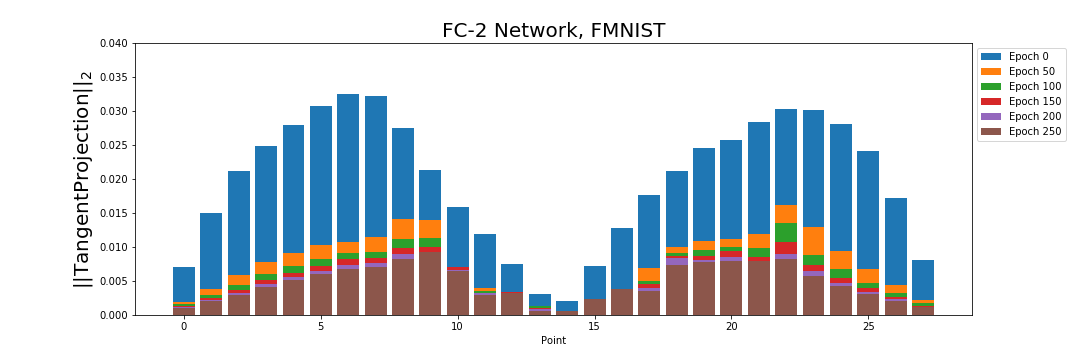}
  \caption{Norm of tangent projection during optimization.}
  \label{fig:exp2}
\end{figure}
\textbf{Optimized paths}. In Figure \ref{fig:exp2}, we show the norm of the tangent projection of the gradient at every point of a path, averaged over an epoch at different training moments. During training, we see a rapid drop in the middle.

\begin{figure}
\centering
  \includegraphics[width=0.8\linewidth]{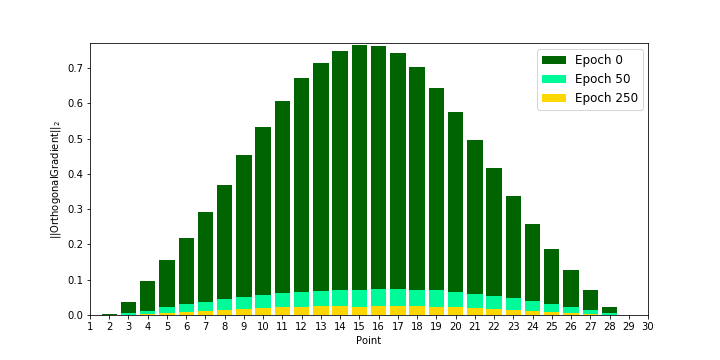}
  \caption{Norm of orthogonal gradient during optimization.}
  \label{fig:orth_grad}
\end{figure}

To illustrate the process of convergence of a single path, in each point, the norm of orthogonal gradient is accumulated after several epochs, see Figure \ref{fig:orth_grad}. We see that at the end of training the values are much lower and nearly zero, telling us that the points of obtained path nearly found flat regions of loss surface and will not change much after optimization. 

\end{document}